\documentclass[10pt,twocolumn,letterpaper]{article}

\usepackage{cvpr}              %

\usepackage[dvipsnames]{xcolor}

\definecolor{cvprblue}{rgb}{0.21,0.49,0.74}
\usepackage[pagebackref,breaklinks,colorlinks,citecolor=cvprblue]{hyperref}
\usepackage{url}
\usepackage[utf8]{inputenc}
\usepackage{booktabs} %
\usepackage{amsfonts} %
\usepackage{nicefrac} %
\usepackage{microtype} %
\usepackage{natbib}
\usepackage{gensymb}
\usepackage{color}
\usepackage{caption}
\usepackage{subcaption}
\usepackage{amsthm}
\usepackage{mathrsfs}
\usepackage{amsmath}
\usepackage{amssymb}
\usepackage{mathtools}
\usepackage{wrapfig}
\usepackage{soul}
\usepackage[ruled,longend, algo2e]{algorithm2e}
\usepackage{diagbox}
\usepackage{algorithm}
\usepackage{algorithmic}
\usepackage{hhline}
\usepackage{multirow}
\usepackage{multicol}
\usepackage{xcolor,colortbl}
\usepackage{soul}

\newcommand{\Mat}{\boldsymbol}

\def\paperID{7663} %
\def\confName{CVPR}
\def\confYear{2024}

\title{ %
Lift3D: Zero-Shot Lifting of Any 2D Vision Model to 3D}

\author{Mukund Varma T\textsuperscript{\textnormal{1}}, Peihao Wang\textsuperscript{\textnormal{2}}, Zhiwen Fan\textsuperscript{\textnormal{2}}, Zhangyang Wang\textsuperscript{\textnormal{2}}, Hao Su\textsuperscript{\textnormal{1}}, Ravi Ramamoorthi\textsuperscript{\textnormal{1}}\\
\textsuperscript{\textnormal{1}}University of California San Diego, \textsuperscript{\textnormal{2}} University of Texas at Austin\\
{\tt\small \{tmukund, haosu, ravir\}@ucsd.edu, \{peihaowang, zhiwenfan, atlaswang\}@utexas.edu}\\
\tt\small \href{https://mukundvarmat.github.io/Lift3D/}{mukundvarmat.github.io/Lift3D/}
}

\newcommand{\ignore}[1]{}

\makeatletter
\renewcommand{\paragraph}{%
  \@startsection{paragraph}{4}%
  {\z@}{0.2ex \@plus 0.3ex \@minus .2ex}{-1em}%
  {\normalfont\normalsize\bfseries}%
}
\makeatother

\begin{document}

\twocolumn[{%
\renewcommand\twocolumn[1][]{#1}%
\maketitle\vspace{-2em}
\includegraphics[width=.99\textwidth]{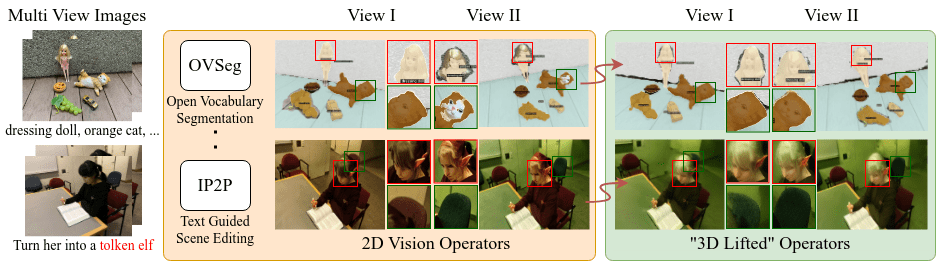}
\captionof{figure}{Imagine we are using a 2D vision operator, such as semantic segmentation or scene editing, on multiple-view input images. This often leads to inconsistent predictions across different views (as shown in the middle column). To address this, we introduce \textit{Lift3D,} a framework designed to transform these inconsistent 2D outputs into view-consistent 3D predictions (illustrated in the right column). Our approach is both scene and operator-agnostic, meaning it can adapt to any downstream task or scene without additional adjustments. We demonstrate how Lift3D effectively resolves inconsistencies in multi-view predictions across open vocabulary segmentation and text-driven scene editing. Notice the color discrepancies in the same rightmost chair across two views (varying from reddish to greenish) in the 2D results at the bottom row, and the inconsistencies in facial and hair color. For a clearer comparison between the 2D and 3D outcomes, we recommend zooming into the electronic version of this image.
\vspace{1em}}
\label{fig:teaser}
}]

\begin{abstract}
In recent years, there has been an explosion of 2D vision models for numerous tasks such as semantic segmentation, style transfer or scene editing, enabled by large-scale 2D image datasets.  At the same time, there has been renewed interest in 3D scene representations such as neural radiance fields from multi-view images.  However, the availability of 3D or multiview data is still substantially limited compared to 2D image datasets, making extending 2D vision models to 3D data highly desirable but also very challenging.  Indeed, extending a single 2D vision operator like scene editing to 3D typically requires a highly creative method specialized to that task and often requires per-scene optimization.  In this paper, we ask the question of whether {\bf any} 2D vision model can be lifted to make 3D consistent predictions.  We answer this question in the affirmative; our new Lift3D method trains to predict unseen views on feature spaces generated by a few visual models (\ie DINO and CLIP), but then generalizes to novel vision operators and tasks, such as style transfer, super-resolution, open vocabulary segmentation and image colorization; for some of these tasks, there is no comparable previous 3D method.  In many cases, we even outperform state-of-the-art methods specialized for the task in question.  Moreover, Lift3D is a zero-shot method, in the sense that it requires no task-specific training, nor scene-specific optimization.  
\vspace{-2em}
\end{abstract}
    
\section{Introduction}
\label{sec:intro}

Recent progress in 2D image understanding has been extraordinary, driven by the assembly of extensive image datasets with intricate labels, and the innovation of varied network architectures.  This has led to remarkable advances in diverse tasks such as semantic segmentation~\cite{kirillov2023segment, liang2023open}, style transfer~\cite{zhang2022domain}, scene editing~\cite{brooks2023instructpix2pix}, and super-resolution~\cite{yue2023resshift}. Yet, the domain of 3D understanding, essential for sectors like autonomous driving, robotics, and 3D asset creation, lags in the development of versatile and robust neural networks. A common requirement for 3D understanding is the processing of multi-view images, which is hindered by the lack of expansive, well-labeled multi-view image datasets. This limitation raises a critical question: Is it possible to modify existing neural networks, initially intended for single-image analysis, to accommodate multi-view inputs, and in doing so, eliminate the inconsistencies typically encountered when applying 2D operators to each view individually?

In this paper, we set an ambitious goal to universally transfer an \textit{arbitrary} pre-trained 2D feature backbone to a 3D model or vision operator \textit{on the fly}, which produces \textit{view-consistent} predictions (see Fig. \ref{fig:teaser}) from any arbitrary viewing angle given a set of posed 2D images and their 2D predictions.
We observe that the intermediate feature maps are roughly aligned with the input image for modern strong 2D operators.
Drawing inspiration from label propagation algorithms \cite{zhu2002learning, zhu2003semi, zhou2003learning, wang2006label, iscen2019label}, for us to produce smooth predictions across views, we only need to rectify inconsistencies and propagate labels from supporting views to novel views.

Motivated by these observations, we propose a novel algorithm {\em Lift3D} which learns to fix and propagate inconsistent multi-view network outputs to view-consistent predictions.
Our architecture design is underpinned by image-based rendering \cite{debevec1996modeling, gortler1996lumigraph, levoy1996light}, where modern methods essentially learn to aggregate pixels with epipolar constraints to synthesize novel views \cite{yu2021pixelnerf, wang2021ibrnet, suhail2021light, suhail2022generalizable, varma2022attention}.
By viewing dense features as colors, our method is trained to interpolate novel views on a feature space generated by a pre-trained 2D visual model or operator, incorporating both RGB and feature maps from adjacent views.

\begin{figure*}[t]
\centering
\includegraphics[width=0.95\linewidth]{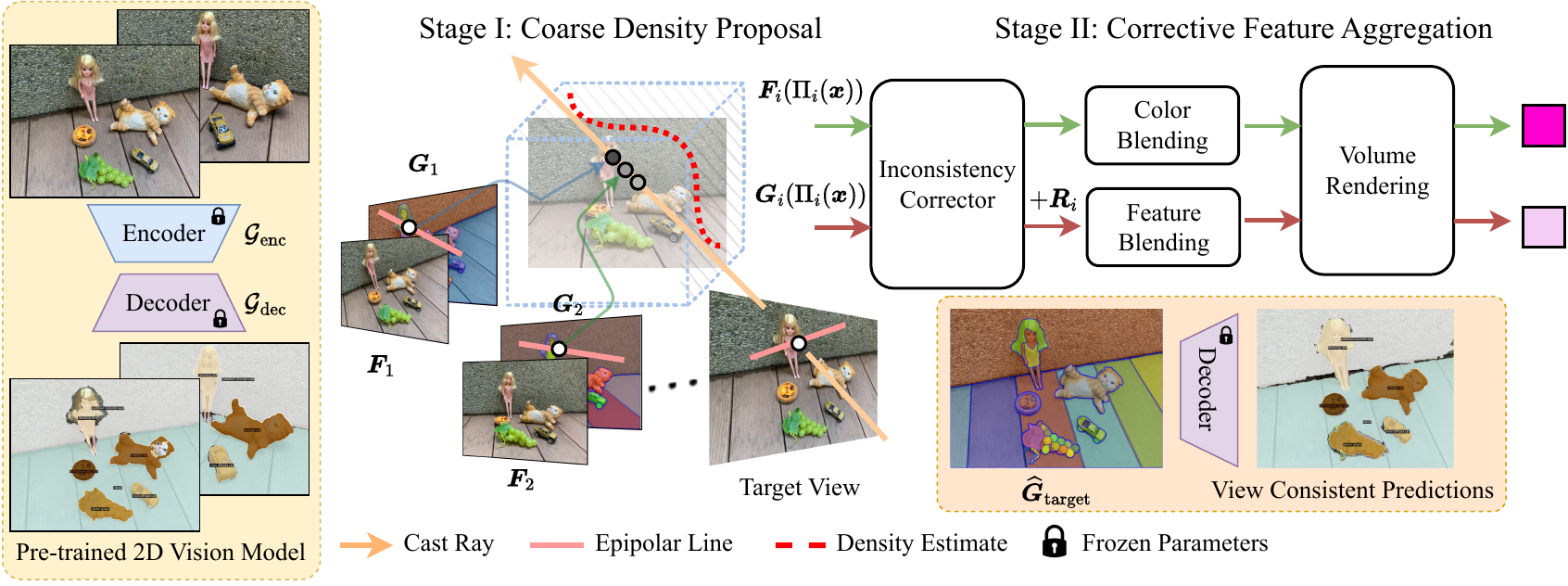}
\caption{Overview of \emph{Lift3D}: 1) Given multi-view images of a scene, we first extract an intermediate feature map using an encoder for each view independently, 2) Using the source view features, we estimate the target view feature map via an extended generalizable novel view synthesis pipeline that learns to correct feature space inconsistency and fuse information according to 3D geometry priors, and 3) Directly use the decoder from the pre-trained visual model to process the estimated feature map and synthesize the final prediction for downstream tasks. }
\label{fig:pipeline}
\end{figure*}

Following the widespread success of Neural Radiance Fields or NeRF for view synthesis~\cite{mildenhall2020nerf, wang2021ibrnet, varma2022attention}, Lift3D casts a ray for each pixel on the target image plane, samples, and projects points to nearby views to fetch RGB and feature values of the epipolar correspondences.
The aggregated RGB information will be used to infer geometry and appearance properties involved in volume rendering \cite{max1995optical, mildenhall2020nerf}.
This geometric information enables the interpolation of features from the pre-trained 2D vision model to estimate a target feature map, that is both multi-view consistent and decodable to generate the desired output (Fig~\ref{fig:pipeline}, Sec.~\ref{sec:method}). 

The whole pipeline is end-to-end differentiable and can be supervised by both ground-truth color and feature maps exported from 2D visual models. An impressive property of Lift3D is that, after we have trained Lift3D on only a few vision operators/models (DINO and CLIP), we discovered that \emph{\textbf{Lift3D has strong zero-shot ability, enabling {\em any} 2D vision operator to be lifted to 3D without any scene-specific or operator-specific training}}. On a variety of 3D vision tasks such as semantic segmentation, style transfer, super-resolution, and text-driven editing of 3D scenes, our performance is comparable and sometimes even better than methods that use per-scene optimization and/or have been designed for the specific task (Fig.~\ref{fig:teaser}, Sec.~\ref{sec:experiments}). In some other tasks like image colorization and open vocabulary segmentation, we present the first 3D extension of 2D vision operators. 

In summary, by formulating feature propagation as a rendering process, Lift3D becomes agnostic to 2D models and task domains, demonstrating versatility to generalize across various feature backbones.
We are unaware of such abilities reported by other works, though there exist several individual works that have dealt with the extension of each separate 2D vision operator to 3D. 
The discovery further supports the practical impact of our method, i.e., as long as there exists a 2D feature backbone for a given task, Lift3D helps realize the same in 3D without any extra fine-tuning.

\section{Related Work}
\label{sec:related}

\paragraph{Progress in 2D Vision Models.} Deep Neural Networks have achieved wide success in a variety of computer vision tasks, ranging from classical object recognition tasks like classification, and segmentation~\cite{dosovitskiy2020image, kirillov2023segment} to more complicated tasks including image generation, editing, etc~\cite{ramesh2021zero, ramesh2022hierarchical, rombach2022high, saharia2022photorealistic}. The 2D image domain boasts of several large, high-quality image datasets~\cite{oquab2023dinov2, schuhmann2022laion, ridnik2021imagenet}.
The recent breakthroughs in natural language processing for large-scale model pre-training have opened the way for similar foundation models in 2D computer vision~\cite{caron2021emerging, oquab2023dinov2, wang2021clip}, that generate visual features at an image and/or pixel level for several downstream applications. 
Given the plethora of 2D models, it is highly desirable to lift these 2D vision operators and tasks to 3D, enabling them to work in a 3D consistent way from multi-view images. 
\paragraph{Attempts in 3D Vision Models.} Some recent efforts \cite{xie2020pointcontrast, huang2023ponder, zhu2023ponderv2} have attempted to build 3D foundational models by mirroring the successful approach of 2D foundation models.
However, this direction is significantly limited by the training data available in 3D.
Another line of work tries to lift a pre-trained 2D foundation model into 3D representations (\eg point clouds or NeRFs~\cite{mildenhall2020nerf}) for 3D scene segmentation~\cite{liu2023segment, liu2023weakly, fan2022nerf} and editing~\cite{fan2022unified, wang2022clip, zhang2022arf, haque2023instruct}.) A major drawback of these approaches is their computational inefficiency as they often involve a per-scene optimization process. Solutions proposed by \cite{ye2023featurenerf, liu2023semantic, chen2023gnesf, chen2023towards} distill a 2D vision model into a generalizable rendering pipeline \cite{wang2021ibrnet, varma2022attention}, that allows for zero-shot inference across different scenes.
Although promising results have been demonstrated on scene segmentation, it remains elusive to generalize these methods to other tasks.
\paragraph{Novel View Synthesis. } A pioneering work, Neural Radiance Fields (NeRF) \cite{mildenhall2020nerf} synthesizes photo-realistic and consistent novel views by fitting each scene as a continuous 5D radiance field (3D coordinates and 2D viewing directions) parameterized by an MLP. Several follow-up works have further improved NeRF's rendering quality~\cite{barron2021mip, barron2022mip, Wizadwongsa2021NeX, wang2021neus}. However, these methods are rarely applied to general 3D applications beyond novel view synthesis, primarily due to the absence of labeled multi-view data. While some methods have been successful in realizing general computer vision tasks in 3D~\cite{wang2022clip, liu2023weakly, haque2023instruct, liu2023stylerf}, they still require a substantial amount of creativity to tailor NeRFs for each task and even need scene-specific optimization.

\section{Preliminary: Generalizable View Synthesis}
\label{sec:gen_nerf}
In this work, we extend the pipeline of Generalizable Novel View Synthesis (GNVS) to render 3D consistent feature maps.
Below we equip readers with the necessary background.
Given \textit{N} calibrated input (or source) views with known pose information $\{\Mat{I}_{i}, \Mat{P}_{i}\}_{i=1}^{N}$, the goal of GNVS is to synthesize target novel view $\Mat{I}_{T}$, even for scenes not observed during training, thereby achieving generalizability.
These methods first extract deep convolutional features $\Mat{F}_{i} = \mathcal{F}_{\text{conv}}(\Mat{I}_{i})$ for each input view. 
To render a target view, several rays are cast into the scene, and $K$ points $\{\Mat{x}_{j}\}_{j=1}^{K}$ as shown in Fig. \ref{fig:pipeline}, are sampled along each ray. 
Each point is then projected onto each source view $\Mat{I}_{i}$ using projection function $\Pi_{i}$, and the nearest feature in the image plane is queried. 
These multi-view features are aggregated into point feature $\Mat{f}(\Mat{x})$ (We drop the index $j$ for simplicity) as follows: 
\begin{align}
\label{eqn:blend}
    \Mat{f}(\Mat{x}) = \mathcal{F}_{\text{view}}(\{\Mat{F}_{i}(\Pi_{i}(\Mat{x}))\}_{i=1}^{N})
\end{align}
where $\Pi_{i}(\Mat{x})$ projects $\Mat{x}$ onto $\Mat{I}_{i}$, $\Mat{F}_{i}(\Pi_{i}(\Mat{x}))$ fetches the feature vector at the epipolar projections for input view $i$, and $\mathcal{F}_{\text{view}}$ is a permutation-invariant aggregation mapping that learns to be occlusion-aware to combine epipolar features~\cite{varma2022attention}. 
In IBRNet~\cite{wang2021ibrnet}, $\mathcal{F}_{\text{view}}$ is implemented by estimating blending weights $\Mat{w}_i$ via a DeepSets-like architecture \cite{zaheer2017deep} to fuse features from different views and produce point feature $\Mat{f}(\Mat{x}) = \sum_i \Mat{w}_i \Mat{F}_{i}(\Pi_{i}(\Mat{x}))$.
Then the alpha value $\Mat{\alpha}(\Mat{x})$ and color $\Mat{c}(\Mat{x})$ can be decoded from feature $\Mat{f}(\Mat{x})$ by another network~\cite{wang2021ibrnet, yu2021pixelnerf}. Finally, using volume rendering~\cite{mildenhall2020nerf}, the point-wise individual color and density values are composed to estimate ray color.

\section{Method} \label{sec:method}
\paragraph{Overview. } We introduce \textit{Lift3D} that applies the effect of any 2D visual models to arbitrary 3D scenes, and illustrate the entire pipeline in Fig. \ref{fig:pipeline}.
Formally, given multi-view images of a scene and a pretrained 2D visual model $\mathcal{G}_{\text{2D}}$, our goal is to generate \textit{view-consistent} predictions from an arbitrary angle, as if the 2D visual model is transferred to be a 3D model.
We assume our 2D visual model follows an encoder-decoder structure\footnote{The 2D visual model need not necessarily follow an encoder-decoder design. We simply refer to the first few layers of the model as $\mathcal{G}_{\text{enc}}$ and the remaining as $\mathcal{G}_{\text{dec}}$.} $\mathcal{G}_{\text{2D}} := \mathcal{G}_{\text{dec}} \circ \mathcal{G}_{\text{enc}}$, where the encoder stage $\mathcal{G}_{\text{enc}}$ maps the input image to a latent representation and the decoder stage $\mathcal{G}_{\text{dec}}$ transforms the latent features to the desired output space. 
We observe that the intermediate latent representation, denoted as $\Mat{G}_{i} = \mathcal{G}_{\text{enc}}(\Mat{I}_{i})$, is robust and spatially aligned with the input image $\Mat{I}_{i}$, despite having inconsistencies across multiple views that need to be fixed.
Our method builds upon existing GNVS techniques~\cite{wang2021ibrnet}, i.e. by viewing dense features as colors, we learn to interpolate novel views on a feature space generated by the 2D pretrained vision model, by incorporating both RGB and feature maps from input views (Sec.~\ref{sec:gen_feat_render}). 
However, naively doing this does not work well since per-view features by 2D models are inherently noisy and inconsistent. 
Rather, our method learns to leverage consistency information on RGB maps to rectify inconsistent artifacts fetched from the feature maps while performing view aggregation (Sec.~\ref{sec:corr_feat_agg}).
By doing so, our method generates the now, view-consistent feature map for the target view that can be directly decoded using $\mathcal{G}_{\text{dec}}$ to synthesize the final prediction i.e. the same task the input pretrained 2D vision model was intended for. 
Such a pipeline can be pre-trained on very few 2D vision models and then directly applied on unseen scenes and 2D vision operators during inference (Sec.~\ref{sec:train_infer}). 
Below, we provide more details on the individual components.

\subsection{Generalizable Feature Rendering}
\label{sec:gen_feat_render}

Existing GNVS techniques~\cite{wang2021ibrnet, suhail2022generalizable, varma2022attention} only render target view color, but it is possible to easily extend them to other quantities of interest, e.g. in this case a high-dimensional feature vector $\Mat{G}_{i}$. 
A straightforward approach is to modify Eq.~\ref{eqn:blend} by replacing $\Mat{F}_{i}(\Pi_{i}(\Mat{x}))$ with $\Mat{G}_{i}(\Pi_{i}(\Mat{x}))$, \ie to fetch the epipolar projections on the per-view intermediate features of the 2D vision model and aggregate them across multiple views to obtain $\Mat{g}(x)$. 
But such a pipeline cannot generalize to any unseen 2D vision operator~\cite{ye2023featurenerf}, as the input intermediate features differ greatly across different 2D vision models. 
Moreover, the absence of ground truth features from arbitrary viewing angles that are multi-view consistent makes training supervision non-trivial. 
However, we have access to RGB information across multiple source views that are view-consistent and can be used to guide the feature correction process.

Motivated by this observation, we simply share the aggregation weights between the epipolar RGB features $\Mat{F}_{i}(\Pi_{i}(\Mat{x}))$ and  similarly constructed epipolar projections on the encoded features from our 2D visual model, denoted by $\Mat{G}_{i}(\Pi_{i}(\Mat{x}))$. 
After we obtain the volumetric representations from RGB features denoted by $\Mat{f}(\Mat{x})$ and from the pretrained vision model $\Mat{g}(\Mat{x})$.
Formally, we rewrite Eq. \ref{eqn:blend} as:
\begin{equation}
\label{eqn:feature_blend}
\begin{aligned}
    \mathcal{F}_{\text{view}} := \Mat{f}(\Mat{x}) = \sum_i \Mat{w}_i \Mat{F}_{i}(\Pi_{i}(\Mat{x}))\\
    \Mat{g}(\Mat{x}) = \sum_i \Mat{w}_i \Mat{G}_{i}(\Pi_{i}(\Mat{x}))\\
    \text{where }\Mat{w}_i = \mathcal{F}_{w}(\{\Mat{F}_{i}(\Pi_{i}(\Mat{x}))\}_{i=1}^{N})
\end{aligned}
\end{equation}
where $\mathcal{F}_{w}$ denotes a fully connected layer that transforms epipolar RGB feature $\Mat{F}_{i}(\Pi_{i}(\Mat{x}))$ to blending weight $\Mat{w}_{i}$.
Next, $\Mat{f}(\Mat{x})$ will be decoded as 
pointwise color and density to predict novel view image $\widehat{\Mat{I}}_{\text{target}}$, while $\Mat{g}(\Mat{x})$ is accumulated along the ray using the estimated density to obtain the feature prediction $\widehat{\Mat{G}}_{\text{target}}$.
In Fig. \ref{fig:pipeline}, we illustrate the epipolar aggregation and ray marching as blending and volume rendering operations respectively. 
By viewing feature rendering as a view interpolation task grounded by RGB input, we ensure generalization to several unseen 2D vision models during inference. 

\subsection{Corrective Feature Aggregation}
\label{sec:corr_feat_agg}

\begin{figure}[t]
\centering
\begin{subfigure}[t]{0.15\textwidth}
  \centering
  \includegraphics[width=1.0\linewidth]{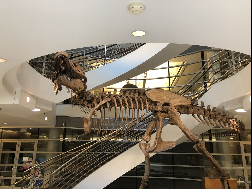}
  \caption{}
\label{fig:baseline1_rgb} 
\end{subfigure}
\begin{subfigure}[t]{0.15\textwidth}
  \centering
  \includegraphics[width=1.0\linewidth]{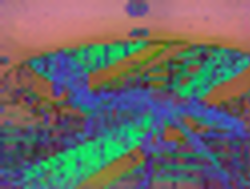}
  \caption{} 
\label{fig:baseline1_est} 
\end{subfigure}
\begin{subfigure}[t]{0.15\textwidth}
  \centering
  \includegraphics[width=1.0\linewidth]{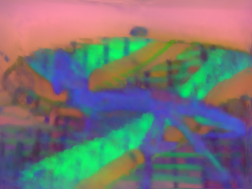}
  \caption{} 
\label{fig:baseline1_gt} 
\end{subfigure}
\caption{(a) Target View, (b) We visualize the PCA (3-channel) map of the interpolated feature maps from our baseline that naively shares blending and ray marching weights between color rendering and 2D encoder features. (c) A similar PCA map is estimated by directly encoding the target view with the 2D encoder. We can see that naively sharing weights without accounting for the inconsistencies in the source view features results in noisy feature estimation, undesirable for several fine-grained downstream tasks. }
\label{fig:baseline1}
\end{figure}

In the previous subsection, we assume the view-consistent RGB feature maps are roughly aligned with the per-view feature derived from the 2D vision model. 
However, in practice this is not the case and several inconsistencies cause significant noise in the rendered feature map. 
In Fig. \ref{fig:baseline1} we visualize a three-channel PCA map of the estimated features after interpolation and compare it against a 2D feature encoding of the target view.
We can clearly see that the estimated novel view features are noisy, therefore inconsistent and do not accurately represent the scene geometry, making them undesirable for several downstream tasks.

As a solution, we propose a two-stage aggregation strategy that first performs a correction on the epipolar features obtained from the 2D visual model, before the remaining aggregation steps. 
The correction factor can be obtained from the RGB features, particularly near the points sampled around the actual 3D location of the object intersecting the cast ray, where multi-view consistency of the epipolar projections is expected. 
We incorporate a coarse-fine stage like NeRF~\cite{mildenhall2020nerf} i.e. perform view synthesis~\cite{wang2021ibrnet} using points randomly sampled along the cast ray, and then perform importance sampling using the obtained coarse density function to specifically fetch projections of points near the actual 3D location of the object of interest. 
The per-view projections on both the RGB and encoder latent feature planes are roughly expected to be view-consistent since they lie on the same object, and can be used for the feature-lifting process.
Therefore, we compute a residual correction term for each view $\Mat{R}_{i}$ as:
\begin{equation}
\centering
\begin{aligned}
    \Mat{\Delta}_{i} = \Mat{F}_{i}(\Pi_{i}(\Mat{x})) - \mathcal{F}_{1}(\Mat{G}_{i}(\Pi_{i}(\Mat{x}))) \\ 
    \Mat{R}_{i} = \mathcal{F}_{2}(\textit{pool}_{i}(\{\Mat{G}_{i}(\Pi_{i}(\Mat{x}))\}_{i=1}^{N}) \oplus \Mat{\Delta}_{i}) \\
    \widetilde{\Mat{G}}_{i}(\Pi_{i}(\Mat{x})) = \Mat{G}_{i}(\Pi_{i}(\Mat{x})) + \Mat{R}_{i}
\end{aligned}
\end{equation}
where $\Mat{x}$ denotes importance sampled points, $\mathcal{F}_{1}$ projects $\Mat{G}_{i}(\Pi_{i}(\Mat{x}))$ into the same latent dimension as $\Mat{F}_{i}(\Pi_{i}(\Mat{x}))$, $\oplus$ denotes the concatenation operation, $\textit{pool}_{i}$ is a max pooling operation applied across input views and $\mathcal{F}_{2}$ represents a fully connected layer that unprojects back to the original dimension size. 
Using these now corrected features, we continue with the aggregation steps detailed in Eq. \ref{eqn:feature_blend} by replacing $\Mat{G}_{i}(\Pi_{i}(\Mat{x}))$ with $\widetilde{\Mat{G}}_{i}(\Pi_{i}(\Mat{x}))$ to synthesize novel view feature (see Fig. \ref{fig:pipeline}).
We ablate on the individual components of our method and include a discussion in the supplementary (see Sec. \ref{sec:abl}).

\subsection{Training and Inference}
\label{sec:train_infer}
The entire pipeline (both the coarse density proposal, and feature renderer networks) can be end-to-end supervised across multiple scenes with their multi-view images and the affiliated 2D encoded feature maps.
\begin{align}
    \mathcal{L} = \lVert \widehat{\Mat{I}}_{\text{target}} - \Mat{I}_{\text{target}} \rVert_2^{2} + \lVert \widehat{\Mat{G}}_{\text{target}} - \Mat{G}_{\text{target}} \rVert_2^{2}
\end{align}
where $\Mat{I}_{\text{target}}$ and $\Mat{G}_{\text{target}}$ denote the target ground truth view and its corresponding feature map encoded by a 2D vision model that is not strictly view-consistent.
However, since the underlying scene geometry is constrained by view-consistent RGB input views, we expect similar propagation in the rendered feature map from the target view. 
Moreover, the simple training objective can cancel out random inconsistent patterns, leading to smooth and precise prediction, with a mechanism similar to \cite{lehtinen2018noise2noise}.
More significantly, we find that our model can be pre-trained with only very few (in practice 2-3) 2D visual models and then directly applied to unseen scenes and 2D vision operators during inference.
Given an unseen 2D vision model that needs to be lifted to 3D, we choose an appropriate intermediate layer and set $\mathcal{G}_{\text{enc}}$, $\mathcal{G}_{\text{dec}}$ based on the choice. 
During inference, given an unseen 2D vision model that needs to be lifted to 3D, we choose an appropriate intermediate layer and set $\mathcal{G}_{\text{enc}}$, $\mathcal{G}_{\text{dec}}$ based on the choice. 
Next, the input views are encoded using $\mathcal{G}_{\text{enc}}$, and a target feature map from any arbitrary view is rendered $\widehat{\Mat{G}}_{\text{target}}$. 
Finally, the desired view-consistent prediction is decoded as $\mathcal{G}_{\text{dec}}(\widehat{\Mat{G}}_{\text{target}})$.
Our model can generalize out of the box since it simplifies 3D feature prediction to a simple interpolation task, which is easy to learn. 
Such interpolation schemes also support truncated or padded features when coming across dimension mismatch with our pre-trained model.
We acknowledge that a view interpolation technique like ours can lead to loss in visual quality, arguably very insignificant (see Figures \ref{fig:teaser}, \ref{fig:gallery}). 
However, in several tasks multi-view consistency is more desirable and repurposing 2D models is a promising direction atleast until the data scarcity problem in 3D is solved.

\section{Results}
\label{sec:experiments}
We conduct experiments to compare {\em Lift3D} against state-of-the-art methods for 3D semantic segmentation, style transfer and scene editing. Since our method is agnostic to 2D models and task domains, we further go on to show results on several other tasks (some not looked at previously in the 3D domain) to demonstrate versatility to generalize across various feature backbones. For all the experiments listed below, our method is never fine-tuned to each scene or trained on the target task or the target feature encoder's features.

\subsection{Implementation Details}
\label{sec:impl}

We train our entire %
pipeline end-to-end on datasets of multi-view posed images. More specifically, we use the training data from IBRNet~\cite{wang2021ibrnet} consisting of synthetic~\cite{downs2022google} and real data~\cite{zhou2018stereo, flynn2019deepview, wang2021ibrnet, mildenhall2021nerf}. During training, we randomly sample $N \in (8, 12)$ source views from a pool of $k \times N$ (where $k \in (1, 3)$) nearby views from the target view. This sampling strategy simulates varying view densities and therefore helps the network generalize better~\cite{wang2021ibrnet}. At every training iteration, we randomly choose a 2D vision model, one of DINO~\cite{caron2021emerging} (ViT-B/8) and CLIP~\cite{wang2021clip} (ViT-B/16) to encode the source and target views. Our method is jointly optimized for view consistent RGB and feature interpolation using the Adam optimizer with an initial learning rate of $5 \times 10^{-4}$, decayed over the course of $250,000$ training steps. We sample $2048$ rays at every iteration, further sampled into $64$ coarse and $128$ fine points.
During inference, we simply apply our lifting process on any given unseen 2D vision model and for any incoming scene without additional optimization.

\subsection{Semantic Segmentation}
\paragraph{Task Description. } Recent methods such as DFF-DINO~\cite{kobayashi2022decomposing}, N3F~\cite{tschernezki2022neural} and ISRF~\cite{goel2023interactive} propose to distill semantic information extracted from 2D models, for example, DINO~\cite{caron2021emerging} onto the 3D feature volume that can be effectively queried (for example using user strokes) to segment objects present in the scene. Specifically, the user provides positive strokes over a region of interest in one view that can be used to determine a high-confidence \textit{seed region} of the object to the segmented. Further, feature matching is performed to match the marked features with the distilled semantics across multiple views. In the context of our method, we lift features from a 2D DINO model (specifically the VIT-S/8 variant) and utilize the same feature-matching strategy used in ~\cite{goel2023interactive} to segment the desired region. We follow the experiment protocol in ISRF to evaluate the performance of our method. 

\begin{table}
  \centering
  \resizebox{\columnwidth}{!}{
  \begin{tabular}{lcccccccc}
    \toprule
    \multirow{2}{*}{Models} & \multicolumn{2}{c}{Chess Table}  & \multicolumn{2}{c}{Color Fountain}  & \multicolumn{2}{c}{Stove} & \multicolumn{2}{c}{Shoe Rack}\\
    \cmidrule(r){2-9}
    & IoU$\uparrow$ & mAP$\uparrow$ & IoU$\uparrow$ & mAP$\uparrow$ & IoU$\uparrow$ & mAP$\uparrow$ & IoU$\uparrow$ & mAP$\uparrow$\\
    \midrule
    N3F \cite{tschernezki2022neural} & 0.344 & 0.334 & 0.871 &  0.871 & 0.416 & 0.387 & 0.589 & 0.582\\
    ISRF \cite{goel2023interactive} & \cellcolor{black!30}0.912 & \cellcolor{black!30}0.916 & \cellcolor{black!15}0.927 & \cellcolor{black!15}0.927 & \cellcolor{black!30}0.819 & \cellcolor{black!30}0.817 & \cellcolor{black!15}0.861 & \cellcolor{black!15}0.869\\
    \midrule
    Lift3D & \cellcolor{black!15}0.824 & \cellcolor{black!15}0.83 & \cellcolor{black!30}0.935 & \cellcolor{black!30}0.938 & \cellcolor{black!15}0.817 & \cellcolor{black!15}0.814 & \cellcolor{black!30}0.871 & \cellcolor{black!30}0.875\\
    \bottomrule
  \end{tabular}}
  \caption{Quantitative results for semantic segmentation using user-guided strokes. We report average scores across the validation views for each scene prescribed by ISRF~\cite{goel2023interactive}. The \sethlcolor{black!30}\hl{best} scores and \sethlcolor{black!15}\hl{second best} scores are highlighted with their respective colors.}
  \label{tab:segm_stroke}
\end{table}

\begin{figure*}[t]
\centering
\begin{subfigure}[t]{0.19\textwidth}
  \centering
  \includegraphics[width=1.0\linewidth]{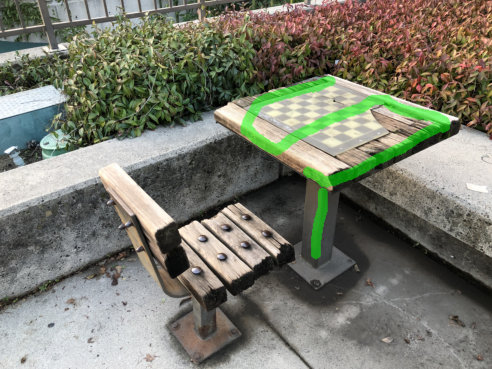}
\label{fig:chess_table_stroke} 
\end{subfigure}%
\begin{subfigure}[t]{0.19\textwidth}
  \centering
  \includegraphics[width=1.0\linewidth]{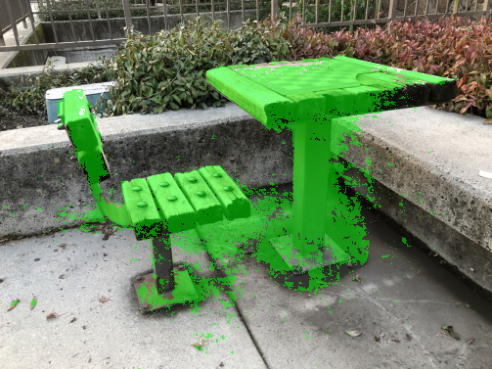}
\label{fig:chess_table_n3f} 
\end{subfigure}%
\begin{subfigure}[t]{0.19\textwidth}
  \centering
  \includegraphics[width=1.0\linewidth]{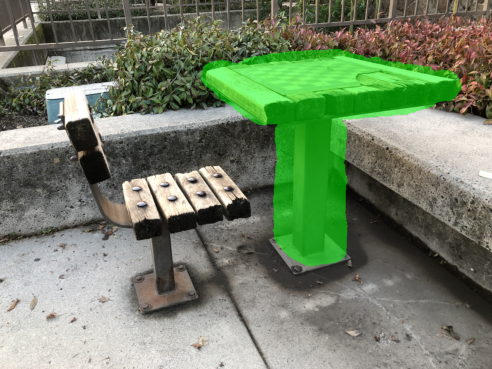}
\label{fig:chess_table_isrf} 
\end{subfigure}%
\begin{subfigure}[t]{0.19\textwidth}
  \centering
  \includegraphics[width=1.0\linewidth]{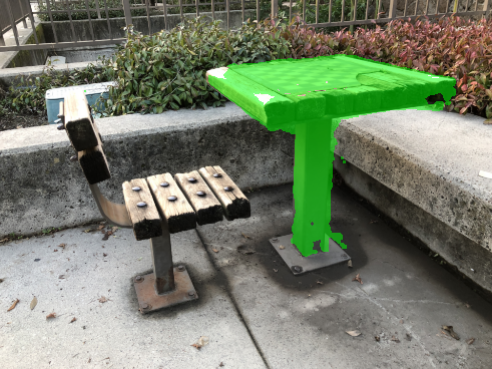}
\label{fig:chess_table_ours} 
\end{subfigure}%
\begin{subfigure}[t]{0.19\textwidth}
  \centering
  \includegraphics[width=1.0\linewidth]{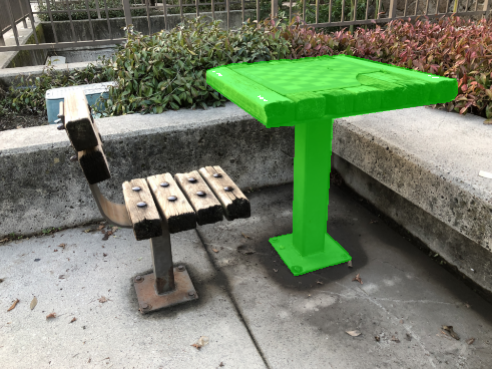}
\label{fig:chess_table_gt} 
\end{subfigure}%
\vspace{-0.5em}
\setcounter{subfigure}{0}
\begin{subfigure}[t]{0.19\textwidth}
  \centering
  \includegraphics[width=1.0\linewidth]{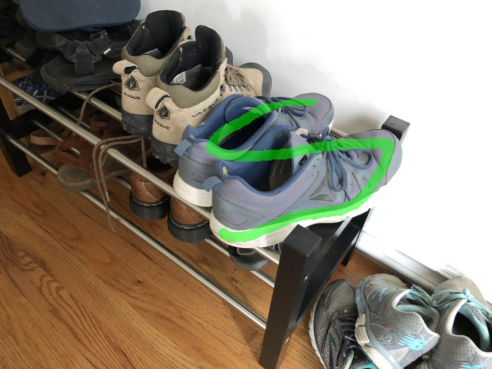}
  \caption{User Stroke}
\label{fig:chess_table_stroke} 
\end{subfigure}%
\begin{subfigure}[t]{0.19\textwidth}
  \centering
  \includegraphics[width=1.0\linewidth]{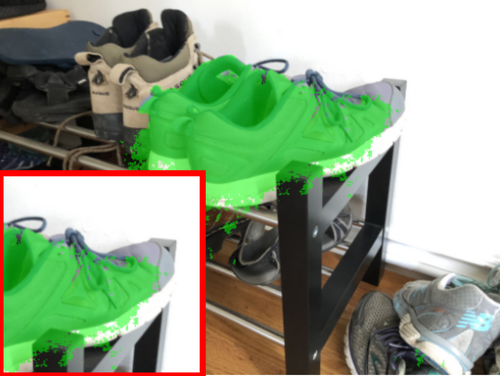}
  \caption{N3F}
\label{fig:chess_table_n3f} 
\end{subfigure}%
\begin{subfigure}[t]{0.19\textwidth}
  \centering
  \includegraphics[width=1.0\linewidth]{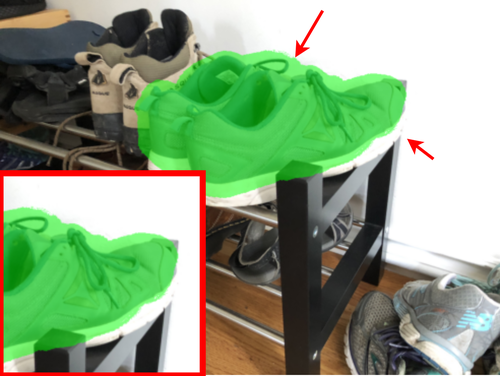}
  \caption{ISRF}
\label{fig:chess_table_isrf} 
\end{subfigure}%
\begin{subfigure}[t]{0.19\textwidth}
  \centering
  \includegraphics[width=1.0\linewidth]{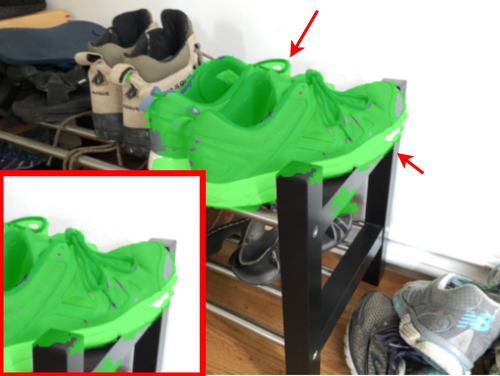}
  \caption{Ours}
\label{fig:chess_table_ours} 
\end{subfigure}%
\begin{subfigure}[t]{0.19\textwidth}
  \centering
  \includegraphics[width=1.0\linewidth]{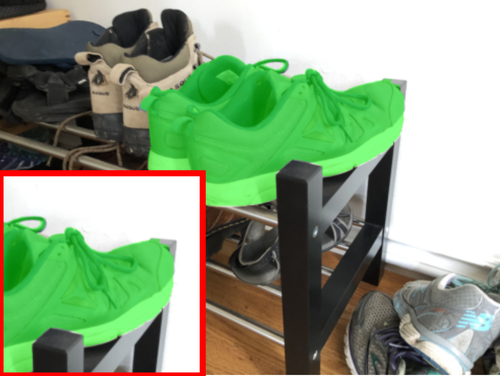}
  \caption{Ground Truth}
\label{fig:chess_table_gt} 
\end{subfigure}%
\caption{Qualitative results for semantic segmentation using user-provided stroke input. In the chess table scene (row 1), our method can derive fine-grained semantic masks, especially around the stem of the table. In the shoe rack scene (row 2), our method can successfully discover the shoe laces and shoe sole better than other methods. }
\label{fig:segm_stroke}
\end{figure*}

\paragraph{Discussion. } We compare our method against recent scene-specific methods N3F~\cite{tschernezki2022neural}, and ISRF~\cite{goel2023interactive}
Table \ref{tab:segm_stroke} presents the segmentation metrics computed over the validation views of each of the 4 scenes presented in ISRF~\cite{goel2023interactive}. It is important to note here that we lift a different variant of the 2D DINO encoder (VIT-S/8) unseen during training to further enforce our \textit{"universality"} claim, while other methods use the much larger variant trained on 768 feature dimensions. Inspite of these differences, our method performs as well as other methods and sometimes even outperforms them on specific scenes. 
This indicates that directly lifting the 2D features to a 3D volume instead of 2D -- 3D feature distillation ensures better retention of the original 2D feature backbone's capabilities. 
In Fig. \ref{fig:segm_stroke}, we visualize these results qualitatively and we can clearly see that our method shows superior segmentation qualitatively (\eg near the shoe laces in Fig.  \ref{fig:segm_stroke}). Even in worse-performing scenes by IoU and mAP (e.g. chesstable), we argue that our method segments clearer boundaries (near the stem of the chess table in Fig. \ref{fig:segm_stroke}) and perhaps the lower metrics are due to the missing base not included in the user stroke leading to ambiguity.

\subsection{Style Transfer}
\paragraph{Task Description. } Given multi-view images of a 3D scene and an image capturing a target style, 3D style transfer aims to generate novel views of the 3D scene that have the target style consistently across the generated views. Naively combining 3D novel view synthesis and 2D style transfer often leads to multi-view inconsistency or poor stylization quality. However, our method Lift3D can transfer a pre-trained 2D feature backbone (SOTA 2D style transfer backbone CAST~\cite{zhang2022domain}) to a 3D model to produce view-consistent predictions from arbitrary viewing angles. To evaluate our method on this task, we compare against the classical image style transfer method AdaIN~\cite{huang2017arbitrary}, SOTA video style transfer methods CCPL~\cite{wu2022ccpl}, ReReVST~\cite{ReReVST2020} and more recent 3D style transfer methods Hyper~\cite{chiang2022stylizing}, LSNV~\cite{huang_2021_3d_scene_stylization} and StyleRF~\cite{liu2023stylerf}. 

\begin{table}
  \centering
  \resizebox{\columnwidth}{!}{
  \begin{tabular}{lcccc}
    \toprule
    \multirow{2}{*}{Models} & \multicolumn{2}{c}{Short-Range Consistency}  & \multicolumn{2}{c}{Long-Range Consistency}\\
    \cmidrule(r){2-5}
    & LPIPS$\downarrow$ & RMSE$\downarrow$ & LPIPS$\downarrow$ & RMSE$\downarrow$\\
    \midrule
    AdaIN \cite{huang2017arbitrary} & 0.279 & 0.085 & 0.452 &  0.167\\
    CCPL \cite{wu2022ccpl} & 0.250 & 0.080 & 0.414 & 0.167\\
    ReReVST \cite{ReReVST2020} & 0.208 & 0.071 & 0.379 & 0.160\\
    LSNV \cite{huang_2021_3d_scene_stylization} & 0.191 & 0.075 & 0.344 & 0.163\\
    Hyper \cite{chiang2022stylizing} & \cellcolor{black!15}0.158 & \cellcolor{black!15}0.066 & \cellcolor{black!30}0.310 & \cellcolor{black!15}0.145\\
    StyleRF \cite{liu2023stylerf} & \cellcolor{black!30}0.138 & 0.071 & 0.333 & 0.165\\
    \midrule
    Lift3D & 0.181 & \cellcolor{black!30}0.053 & \cellcolor{black!15}0.327 & \cellcolor{black!30}0.138\\
    \bottomrule
  \end{tabular}}
  \caption{Quantitative results for 3D scene style transfer. We compute the short and long-term consistency metrics across all 8 scenes from the LLFF dataset. }
  \label{tab:style_transfer}
\end{table}

\begin{figure*}[t]
\centering
\begin{subfigure}[t]{0.19\textwidth}
  \centering
  \includegraphics[width=1.0\linewidth]{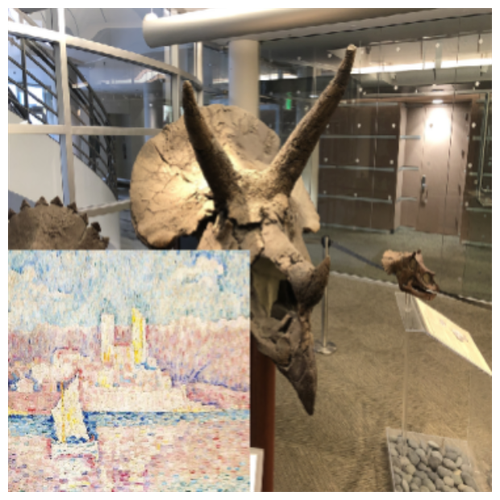}
\label{fig:horns1_hyper_view1} 
\end{subfigure}%
\begin{subfigure}[t]{0.19\textwidth}
  \centering
  \includegraphics[width=1.0\linewidth]{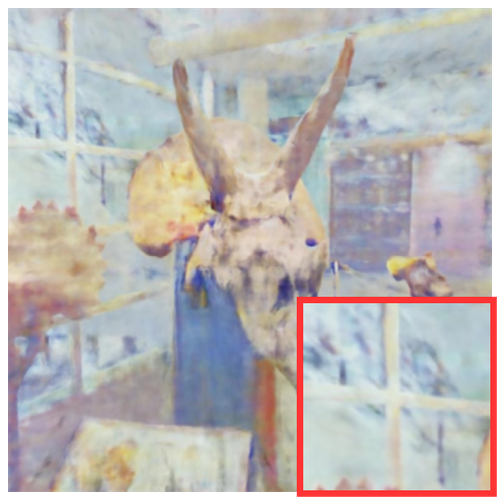}
\label{fig:horns1_hyper_view1} 
\end{subfigure}%
\begin{subfigure}[t]{0.19\textwidth}
  \centering
  \includegraphics[width=1.0\linewidth]{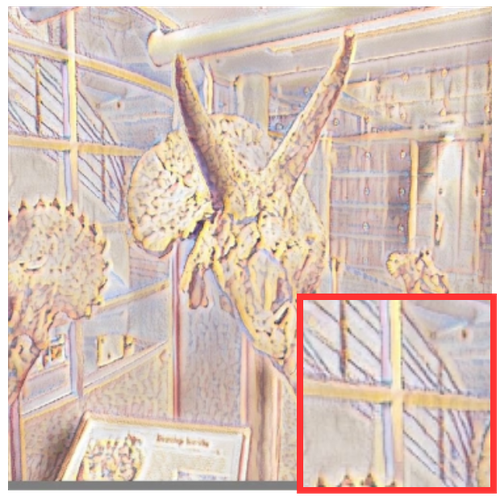}
\label{fig:horns1_lsnv_view1} 
\end{subfigure}%
\begin{subfigure}[t]{0.19\textwidth}
  \centering
  \includegraphics[width=1.0\linewidth]{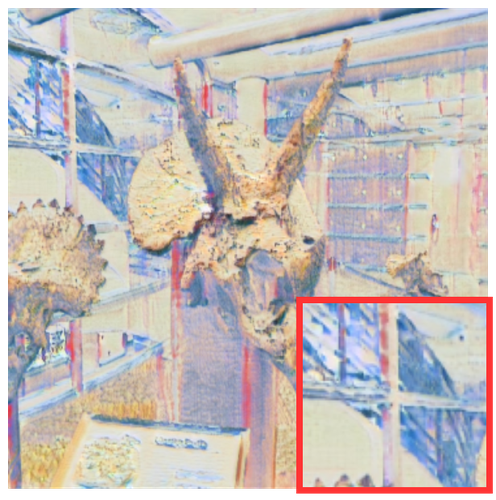}
\label{fig:horns1_stylerf_view1} 
\end{subfigure}%
\begin{subfigure}[t]{0.19\textwidth}
  \centering
  \includegraphics[width=1.0\linewidth]{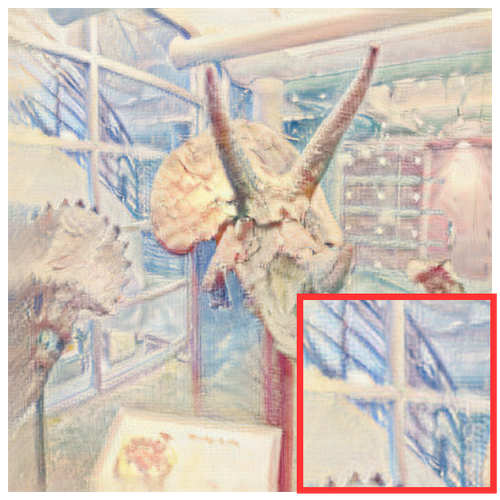}
\label{fig:horns1_ours_view1} 
\end{subfigure}%
\vspace{-1em}
\setcounter{subfigure}{0}
\begin{subfigure}[t]{0.19\textwidth}
  \centering
  \includegraphics[width=1.0\linewidth]{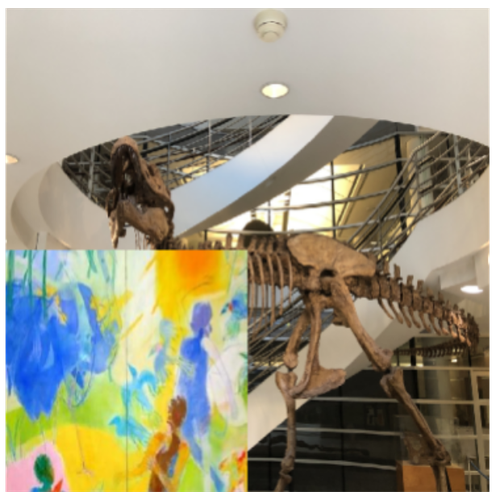}
  \caption{Original / Style}
\label{fig:trex1_hyper_view1} 
\end{subfigure}%
\begin{subfigure}[t]{0.19\textwidth}
  \centering
  \includegraphics[width=1.0\linewidth]{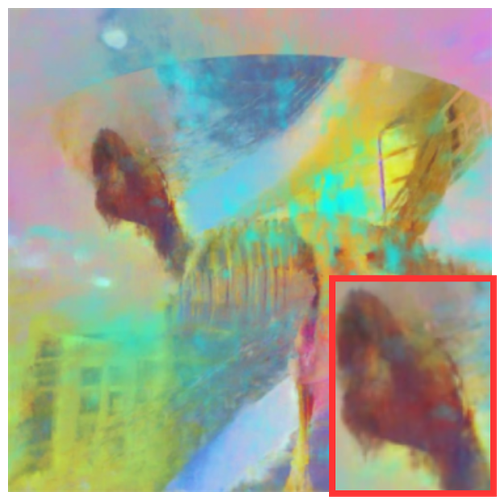}
  \caption{Hyper}
\label{fig:trex1_hyper_view1} 
\end{subfigure}%
\begin{subfigure}[t]{0.19\textwidth}
  \centering
  \includegraphics[width=1.0\linewidth]{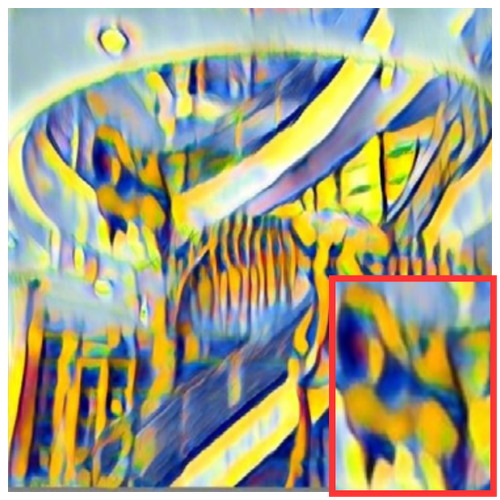}
  \caption{LSNV}
\label{fig:trex1_lsnv_view1} 
\end{subfigure}%
\begin{subfigure}[t]{0.19\textwidth}
  \centering
  \includegraphics[width=1.0\linewidth]{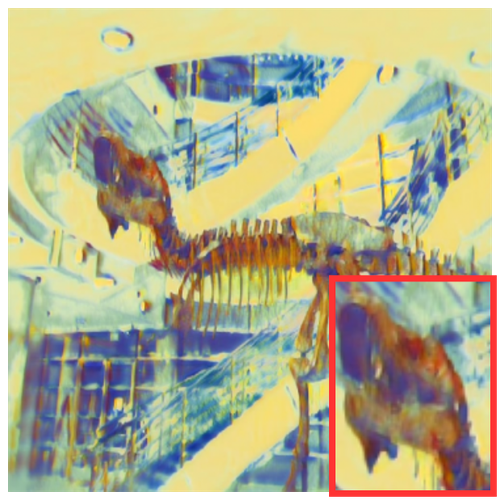}
  \caption{StyleRF}
\label{fig:trex1_stylerf_view1} 
\end{subfigure}%
\begin{subfigure}[t]{0.19\textwidth}
  \centering
  \includegraphics[width=1.0\linewidth]{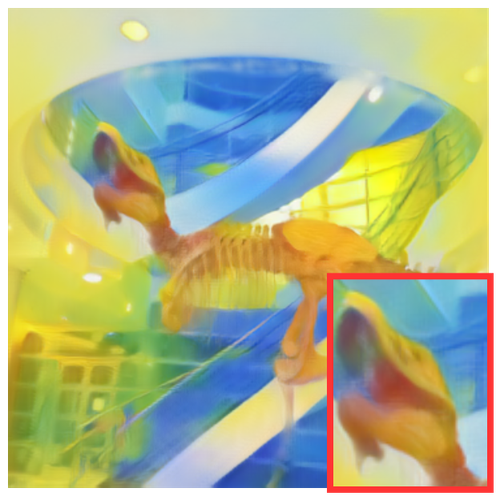}
  \caption{Ours}
\label{fig:trex1_ours_view1} 
\end{subfigure}%
\caption{Qualitative results for 3D scene style transfer. Our method achieves better retention of the original geometry (row 1) and is more coherent to the style image (row 2). }
\label{fig:style_transfer}
\end{figure*}

\paragraph{Discussion. } We evaluate our method over the LLFF dataset~\cite{mildenhall2019local}, containing real forward-facing scenes with complex geometry structures. Following the protocol in StyleRF, we measure the multi-view stylization consistency by warping one view to the other based on optical flow~\cite{teed2020raft} and then compute the masked RMSE and LPIPS score~\cite{zhang2018perceptual} across nearby views (short-range consistency) and far-away views (long-range consistency). It can be seen from Table \ref{tab:style_transfer} that our method achieves the best RMSE score and competitive LPIPS scores when compared to other methods. Please note that these metrics do not adequately measure the stylization quality and over-smoothened results (e.g. in the case of Hyper) tend to produce better scores~\cite{liu2023stylerf}. Fig. \ref{fig:style_transfer} visualizes qualitative results and we can clearly see that our method achieves better stylization (row 2) while retaining the original scene geometry better (row 1). Furthermore, our method Lift3D generalizes to the style transfer task in a zero-shot manner unlike other stylization-specific methods that still require per-scene tuning. 

\begin{figure*}[hbt!]
\centering
\rotatebox[origin=l]{90}{``... car"}%
\rotatebox[origin=l]{90}{``\textcolor{red}{gold} car"}%
\hspace{0.1em}
\begin{subfigure}[t]{0.16\textwidth}
  \centering
  \includegraphics[width=1.0\linewidth]{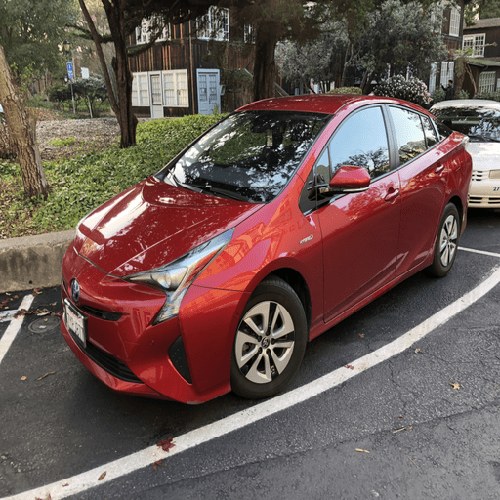}
\label{fig:orig_redtoyota_view1} 
\end{subfigure}%
\begin{subfigure}[t]{0.16\textwidth}
  \centering
  \includegraphics[width=1.0\linewidth]{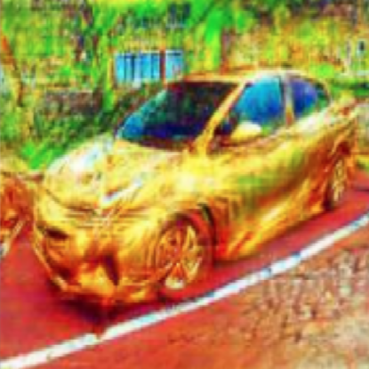}
\label{fig:clipnerf_redtoyota_view1} 
\end{subfigure}%
\begin{subfigure}[t]{0.16\textwidth}
  \centering
  \includegraphics[width=1.0\linewidth]{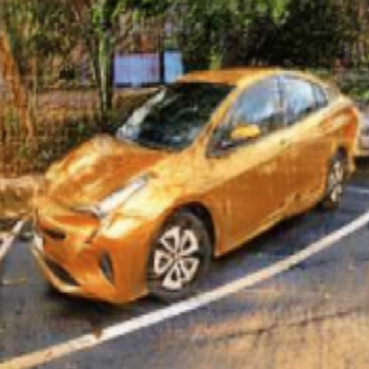}
\label{fig:nerfart_redtoyota_view1} 
\end{subfigure}%
\begin{subfigure}[t]{0.16\textwidth}
  \centering
  \includegraphics[width=1.0\linewidth]{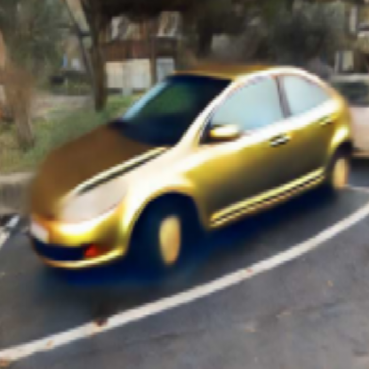}
\label{fig:maskedsds_redtoyota_view1} 
\end{subfigure}%
\begin{subfigure}[t]{0.16\textwidth}
  \centering
  \includegraphics[width=1.0\linewidth]{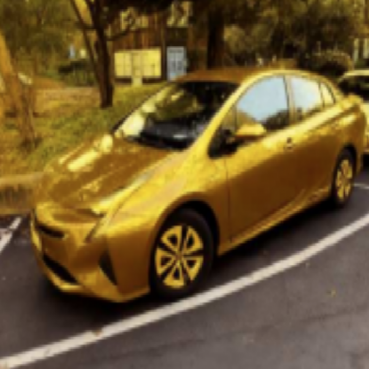}
\label{in2n_redtoyota_view1} 
\end{subfigure}%
\begin{subfigure}[t]{0.16\textwidth}
  \centering
  \includegraphics[width=1.0\linewidth]{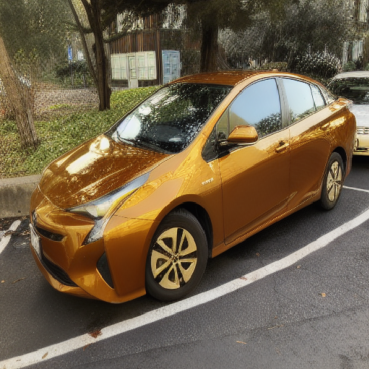}
\label{ours_redtoyota_view1} 
\end{subfigure}%
\newline
\setcounter{subfigure}{0}
\rotatebox[origin=l]{90}{``... flower"}%
\rotatebox[origin=l]{90}{``\textcolor{red}{popcorn} flower"}%
\hspace{0.1em}
\begin{subfigure}[t]{0.16\textwidth}
  \centering
  \includegraphics[width=1.0\linewidth]{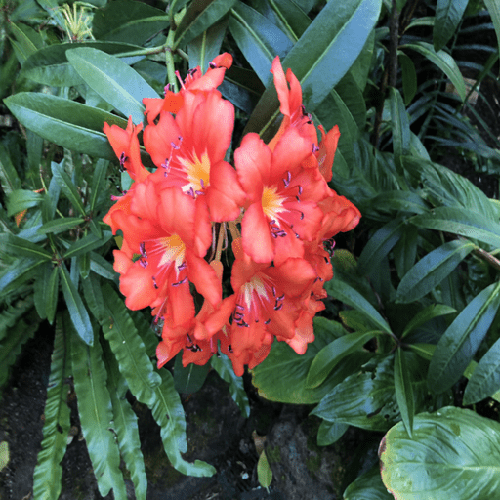}
  \caption{Original}
\label{fig:orig_flower_view1} 
\end{subfigure}%
\begin{subfigure}[t]{0.16\textwidth}
  \centering
  \includegraphics[width=1.0\linewidth]{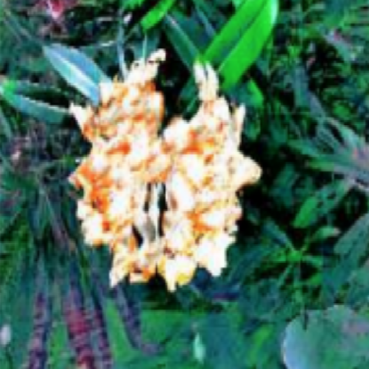}
  \caption{CLIP NeRF}
\label{fig:clipnerf_flower_view1} 
\end{subfigure}%
\begin{subfigure}[t]{0.16\textwidth}
  \centering
  \includegraphics[width=1.0\linewidth]{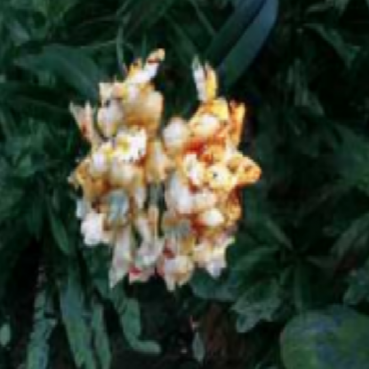}
  \caption{NeRF ART}
\label{fig:nerfart_flower_view1} 
\end{subfigure}%
\begin{subfigure}[t]{0.16\textwidth}
  \centering
  \includegraphics[width=1.0\linewidth]{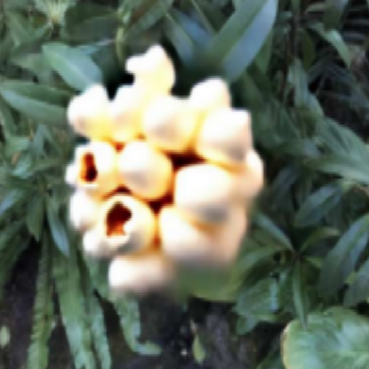}
  \caption{Masked SDS}
\label{fig:maskedsds_flower_view1} 
\end{subfigure}%
\begin{subfigure}[t]{0.16\textwidth}
  \centering
  \includegraphics[width=1.0\linewidth]{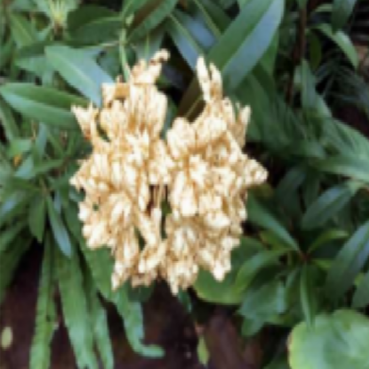}
  \caption{InstructNeRF2NeRF}
\label{fig:in2n_flower_view1} 
\end{subfigure}%
\begin{subfigure}[t]{0.16\textwidth}
  \centering
  \includegraphics[width=1.0\linewidth]{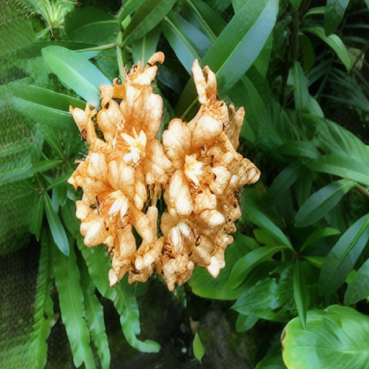}
  \caption{Ours}
\label{fig:ours_flower_view1} 
\end{subfigure}%
\setcounter{subfigure}{6}
\hspace{1em}
\begin{subfigure}[t]{0.16\textwidth}
  \centering
  \includegraphics[width=1.0\linewidth]{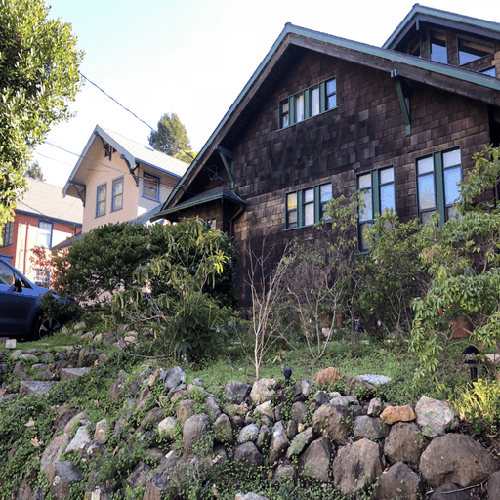}
  \caption{Original \\ ``make it \textcolor{red}{minecraft}"}
\label{fig:house_orig} 
\end{subfigure}%
\begin{subfigure}[t]{0.16\textwidth}
  \centering
  \includegraphics[width=1.0\linewidth]{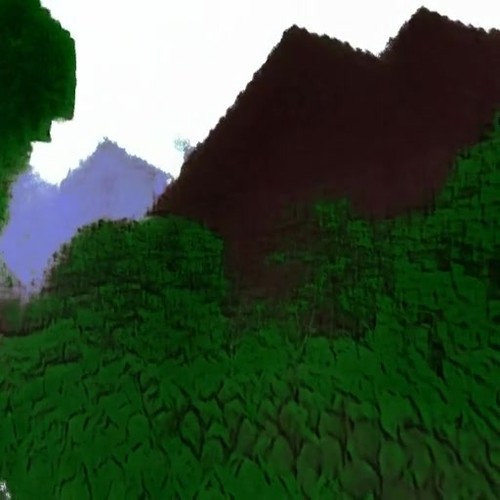}
  \caption{InstructNeRF2NeRF}
\label{fig:house_in2n} 
\end{subfigure}%
\begin{subfigure}[t]{0.16\textwidth}
  \centering
  \includegraphics[width=1.0\linewidth]{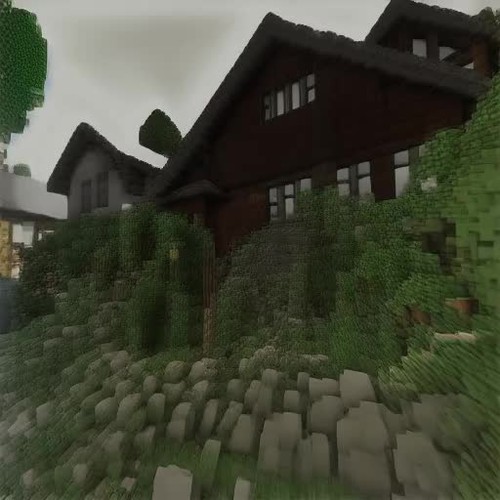}
  \caption{Ours}
\label{fig:house_ours} 
\end{subfigure}
\begin{subfigure}[t]{0.16\textwidth}
  \centering
  \includegraphics[width=1.0\linewidth]{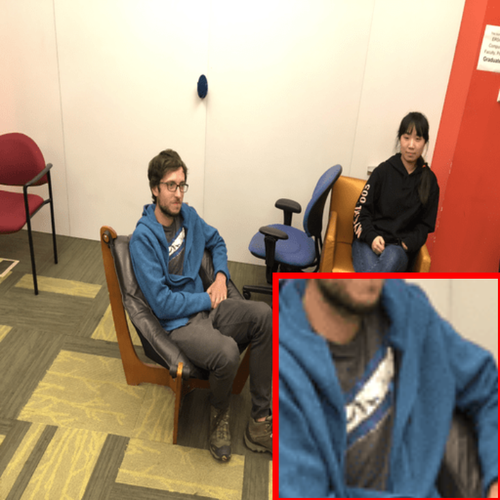}
  \caption{Original \\ ``make them \textcolor{red}{anime}"}
\label{fig:mattcecsit_orig} 
\end{subfigure}%
\begin{subfigure}[t]{0.16\textwidth}
  \centering
  \includegraphics[width=1.0\linewidth]{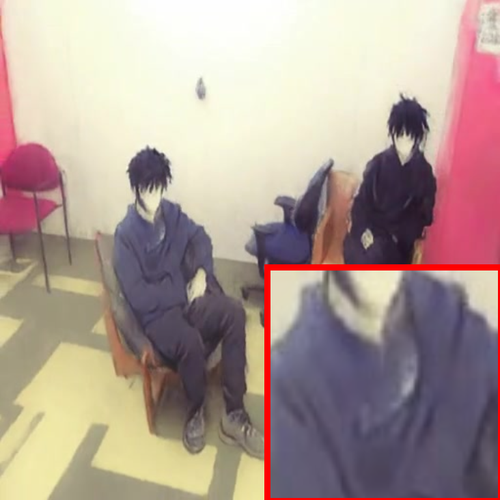}
  \caption{InstructNeRF2NeRF}
\label{fig:mattcecsit_in2n} 
\end{subfigure}%
\begin{subfigure}[t]{0.16\textwidth}
  \centering
  \includegraphics[width=1.0\linewidth]{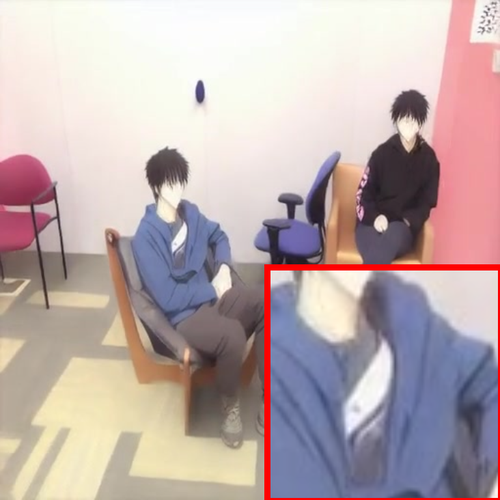}
  \caption{Ours}
\label{fig:mattcecsit_ours} 
\end{subfigure}%
\caption{Qualitative results on the text-driven scene editing task. Our method consistently achieves superior editing quality while still retaining the original scene geometry compared to previous work. In row 3, we can clearly see that our method retains the structure of the house, and clothes worn by the people while still adhering to the text prompt. } 
\label{fig:scene_edit}
\end{figure*}
\subsection{Scene Editing. }
\paragraph{Task Description. } In this task, we look at editing a 3D implicit representation (in this case a NeRF) using text instructions. We attempt to lift a 2D diffusion model InstructPix2Pix~\cite{brooks2023instructpix2pix} to construct a view-consistent feature volume that can be decoded for text-based 3D editing. To evaluate our method on this task, we compare against CLIP-NeRF~\cite{wang2022clip}, NeRF-Art~\cite{wang2023nerf} and Masked SDS~\cite{poole2022dreamfusion} that fine-tune the parameters of a NeRF model using 2D supervision either from a pretrained CLIP or Diffusion model, and more recent methods like Instruct NeRF-to-NeRF~\cite{haque2023instruct} that directly manipulate the training images of NeRF.

\paragraph{Discussion. }
As seen in Fig. \ref{fig:scene_edit}, our method shows the capability to edit scenes with superior quality compared to other methods. Prior work that leverages CLIP-based optimization~\cite{wang2022clip, wang2023nerf} struggles to produce reasonable edited images due to the inherent limitations in the CLIP model. More recent diffusion-based approaches~\cite{poole2022dreamfusion, haque2023instruct} produce higher quality results but either alter the image too severely or too little (see Figs. \ref{fig:maskedsds_flower_view1}, \ref{fig:in2n_flower_view1}), while our method consistently does well across several edits (we present more results in the supplementary). Further, we provide more examples that compare our method against InstructNeRF2NeRF, a recent SOTA method for instruction-based editing of 3D scenes, and leverage the same 2D feature backbone as ours. We can clearly see that our method ensures retention of the original scene geometry (see Figs. \ref{fig:house_ours}, \ref{fig:mattcecsit_ours}) while still adhering to the editing text prompt. These results clearly show that directly lifting the intermediate latent of the 2D diffusion model outperforms naive optimization strategies that edit NeRF's training images. 
We defer quantitative results that measure view-consistency and text-similarity of the edited scenes to the supplementary (see Table \ref{tab:sceneedit}).

\begin{figure*}[hbt!]
\centering
\raisebox{-0.25in}{\rotatebox[origin=t]{90}{Open Vocabulary Segmentation}}%
\hspace{0.1em}
\begin{subfigure}[t]{0.48\textwidth}
  \centering
  \includegraphics[width=0.49\linewidth]{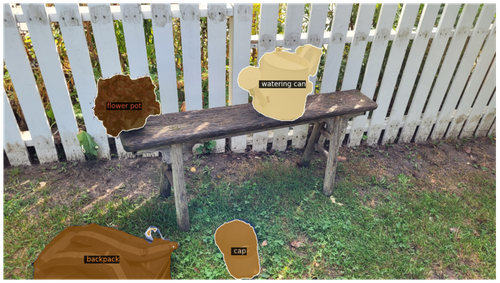}
  \includegraphics[width=0.49\linewidth]{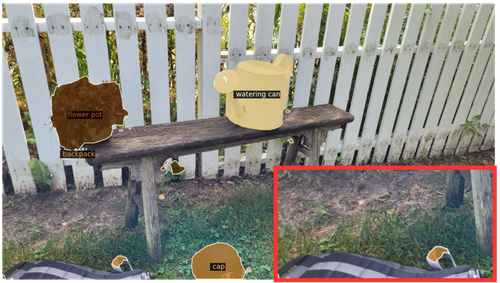}
  \includegraphics[width=0.49\linewidth]{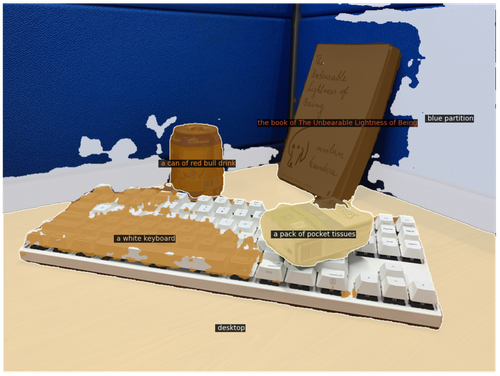}
  \includegraphics[width=0.49\linewidth]{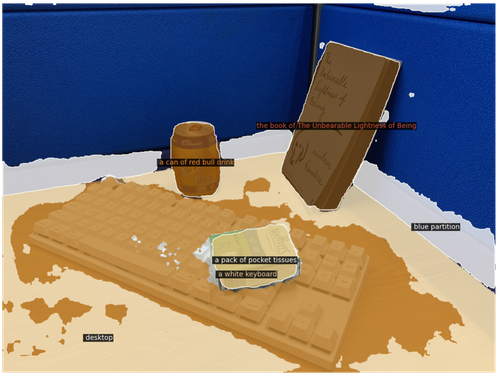}
  \caption{2D OVSeg}
\label{fig:ovseg2d} 
\end{subfigure}%
\hspace{0.2em}
\begin{subfigure}[t]{0.48\textwidth}
  \centering
  \includegraphics[width=0.49\linewidth]{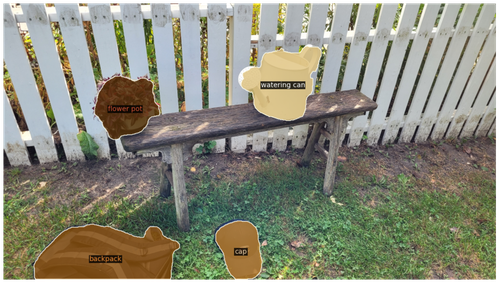}
  \includegraphics[width=0.49\linewidth]{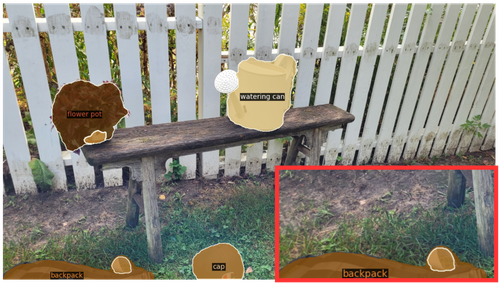}
  \includegraphics[width=0.49\linewidth]{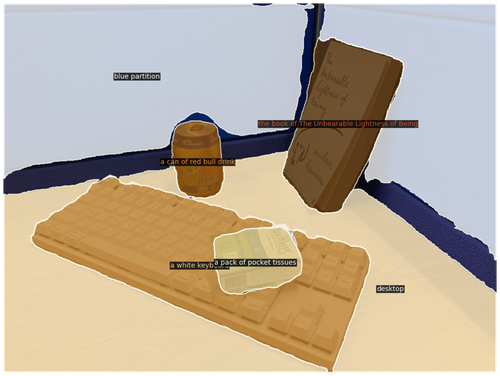}
  \includegraphics[width=0.49\linewidth]{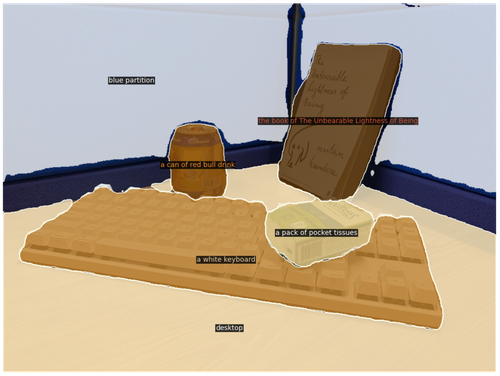}
  \caption{3D ``Lifted" OVSeg}
\label{fig:ovseg3d} 
\end{subfigure}%
\newline
\raisebox{-0.25in}{\rotatebox[origin=t]{90}{Colorization}}%
\hspace{0.1em}
\begin{subfigure}[t]{0.48\textwidth}
  \centering
  \includegraphics[width=0.49\linewidth]{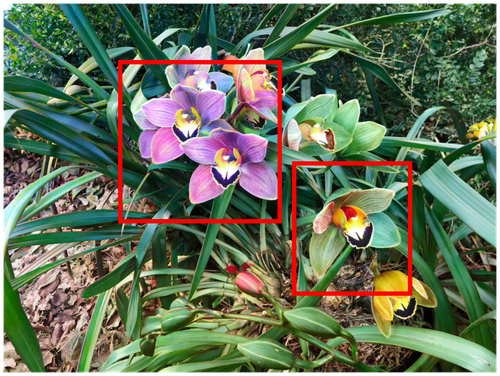}
  \includegraphics[width=0.49\linewidth]{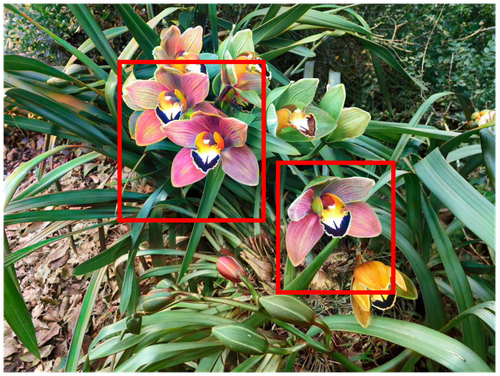}
  \includegraphics[width=0.49\linewidth]{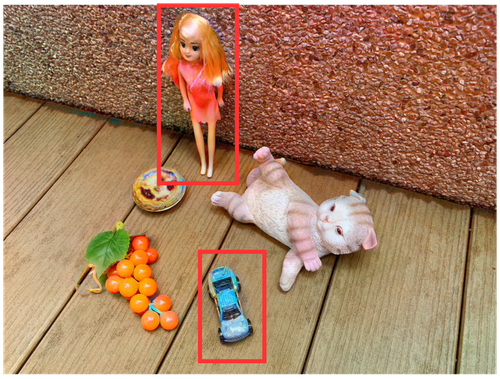}
  \includegraphics[width=0.49\linewidth]{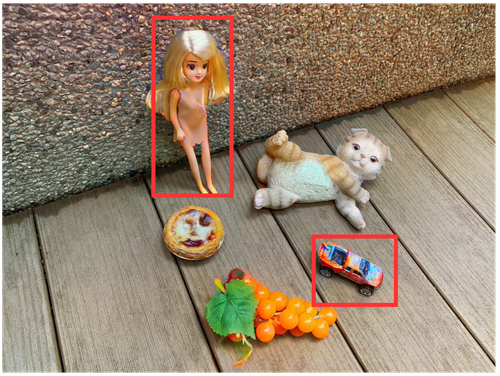}
  \caption{2D DDColor}
\label{fig:ddcolor2d} 
\end{subfigure}%
\hspace{0.2em}
\begin{subfigure}[t]{0.48\textwidth}
  \centering
  \includegraphics[width=0.49\linewidth]{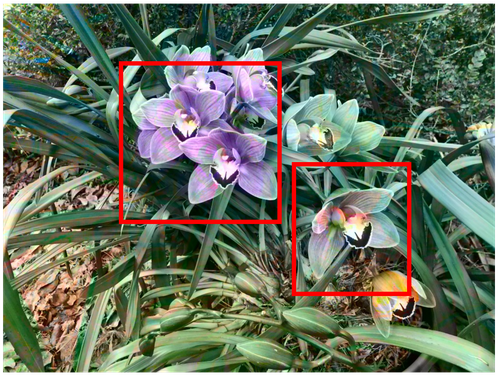}
  \includegraphics[width=0.49\linewidth]{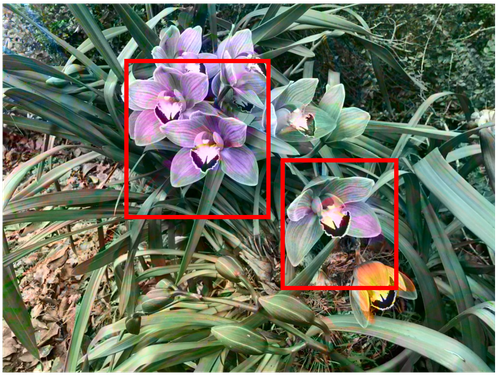}
  \includegraphics[width=0.49\linewidth]{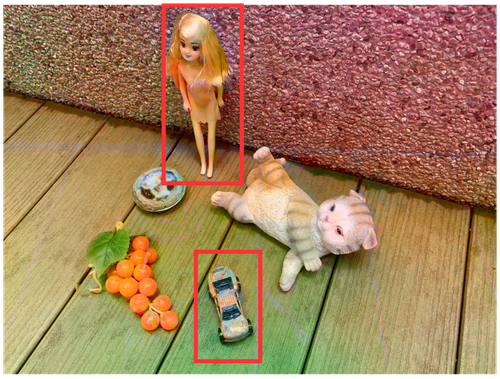}
  \includegraphics[width=0.49\linewidth]{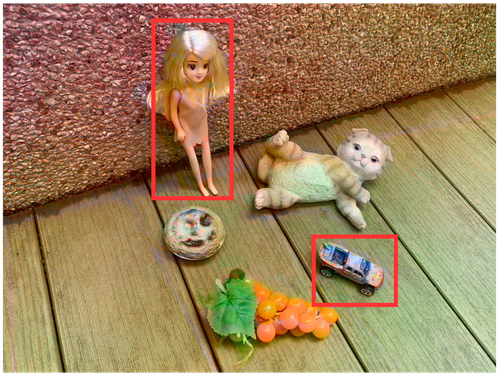}
  \caption{3D ``Lifted" DDColor}
\label{fig:ddcolor3d} 
\end{subfigure}%
\caption{Qualitative results on open vocabulary segmentation and colorization, across two views. Our 3D ``lifted" features can successfully segment occluded objects (row 1) and with superior quality (row 2). Moreover, they also show improved view consistency compared to the 2D operation in both tasks. }
\label{fig:other_tasks}
\end{figure*}
\subsection{Other Tasks}
In the previous subsections, we discussed the generality of our proposed method to lift several 2D feature backbones to generate 3D consistent predictions without any extra training. This enables us to realize tasks that have not been previously looked at in the 3D domain. This underscores the generality of our method i.e. as long as there exists a 2D feature backbone for a given task, Lift3D helps realize the same in 3D without any extra training or efforts on data collection.

\paragraph{Open Vocabulary Segmentation. } Given the multi-view images of a 3D scene and the open-vocabulary text descriptions, we wish to derive accurate object boundaries for the scene. Some recent work~\cite{kerr2023lerf} distills large 2D vision-language models onto a 3D radiance field, which can be effectively used to obtain relevancy maps for language queries. However, the obtained semantic masks are very coarse and cannot be used for localization. Therefore, the task of open vocabulary segmentation in the 3D domain is still an open problem. Towards this end, we transfer the intermediate features of a SOTA 2D open vocabulary segmentation method OVSeg~\cite{liang2023open} to a 3D feature volume to derive multi-view consistent and accurate semantic masks. Figures \ref{fig:ovseg2d} and \ref{fig:ovseg3d} visualize predictions obtained at multiple viewpoints by 1) naively applying 2D OVSeg on multi-view data and 2) using the 3D ``Lifted" OVSeg features estimated by Lift3D. We can clearly see that the lifted features ensure multi-view consistent predictions (also see Fig. \ref{fig:teaser}) and surprisingly even improve segmentation quality (precise keyboard boundaries). By using multi-view information our model can reason about occlusions and other variabilities across arbitrary viewpoints, resulting in enhanced segmentation masks.

\paragraph{Image Colorization. } The image colorization task utilizes grayscale multi-view images to generate plausible colored NeRF scenes. The ambiguity in 2D image colorization leads to severe multi-view inconsistencies and we verify if our proposed strategy can rectify them by adhering to multi-view geometry. We leverage SOTA single view colorization technique DDColor~\cite{kang2022ddcolor} and construct a 3D feature volume that can be decoded to estimate the colored image. Figs. \ref{fig:ddcolor2d} and \ref{fig:ddcolor3d} compare the lifted 3D features against 2D image colorization of each view individually. We can clearly see that our method ensures consistent colors across different views while preserving a similar output quality as the original network. These results clearly verify the adaptability of our proposed pipeline to several tasks.

\section{Conclusion}

We present \emph{Lift3D}, a generalizable system that can lift any 2D visual model to synthesize view-consistent feature predictions without training with data from downstream tasks.
Our method essentially learns to rectify and propagate predicted feature maps of source views to synthesize the feature map for the novel view.
Our algorithm mitigates inconsistencies among source view predictions, and generates view-smooth predictions at the target view. 
We demonstrate that Lift3D is merely pre-trained on DINO and CLIP features but can directly generalize to a wide range of 2D vision models, empowering various applications, including semantic segmentation, stylization, instructed scene editing, and many others.
All empirical observations endorse that Lift3D can be a crucial component in bringing the recent advancement of 2D vision models to the 3D domain.

\vspace{-1em}
\subsection*{Acknowledgements}
This work was funded in part by a Jacobs School of Engineering Fellowship, ONR grant (N00014-23-1-2526), NSF grants (2100237, 2120019), NSF CAREER Award (2240160), NSF TILOS AI Institute (2112665), the Ronald L. Graham Chair and the UC San Diego Center for Visual Computing.  We also acknowledge support from IARPA via contract 140D0423C0076 and gifts from Adobe, Google, Qualcomm and Rembrand. 
\vspace{-3em}

{
    \small
    \bibliographystyle{ieeenat_fullname}
    \bibliography{main}
}

\clearpage
\setcounter{page}{1}
\appendix
\maketitlesupplementary

\section{Ablation Studies}
\label{sec:abl}

\begin{table}
  \centering
  \resizebox{\columnwidth}{!}{
  \begin{tabular}{lccc}
    \toprule
    Model & ResShift~\cite{yue2023resshift} & DDColor~\cite{kang2022ddcolor} & OVSeg~\cite{liang2023open}\\
    \midrule
    Feature Prediction & 3.121 & 12.282 & 10.117\\
    Feature Interpolation &  &  & \\
    \quad w/o Inconsistency Correction & 0.042 & 7.491 & 0.375\\
    \quad w/ Inconsistency Correction &  &  & \\
    \quad\quad Single Stage & \cellcolor{black!15}0.0395 & \cellcolor{black!15}3.39 & \cellcolor{black!15}0.371\\
    \midrule
    \quad\quad Ours & \cellcolor{black!30}0.021 & \cellcolor{black!30}3.240 & \cellcolor{black!30}0.370\\
    \bottomrule
  \end{tabular}}
  \caption{Ablation study of several components of Lift3D on lifting three unseen 2D feature encoders to 3D. The indent indicates the studied setting is added upon the upper-level ones.}
  \label{tab:abl}
\end{table}

We primarily ablate on the following key design considerations of our method. 

\paragraph{Feature Prediction. } Inspired by ~\cite{ye2023featurenerf}, we propose a baseline that directly predicts the target view feature and color. To do so, we leverage a powerful generalizable NVS method GNT~\cite{varma2022attention}, and add an additional feature head that is supervised by the 2D vision model.  

\paragraph{w/o Inconsistency Correction. } Next, We remove the inconsistency correction module on the source view features and simply share the aggregation weights between the epipolar RGB features and similarly constructed epipolar projections on the encoder features from the 2D vision model.  

\paragraph{Single Stage. } Lastly, we convert our two-stage pipeline into a ``one-stage'' feature correction pipeline using the randomly sampled points along the ray (in contrast to our two-stage, density proposal followed by feature propagation only on the importance-sampled points). 

We report performance on the above investigations in Table \ref{tab:abl}, specifically the mean squared error distance between the estimated feature and ground truth feature obtained by naively encoding the target view using the 2D vision model (or $\Mat{G}_{\text{target}}$). We follow the training strategy discussed in Sec. \ref{sec:impl} {\ie trained on DINO and CLIP features and evaluated on unseen 2D vision models - ResShift~\cite{yue2023resshift}, DDColor~\cite{kang2022ddcolor}, and OVSeg~\cite{liang2023open}}. 

\begin{figure}[t]
\centering
\begin{subfigure}[t]{0.23\textwidth}
  \centering
  \includegraphics[width=1.0\linewidth]{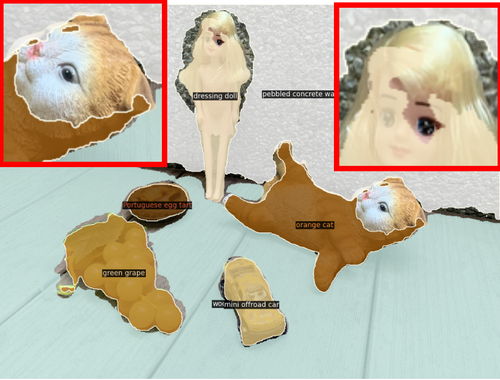}
  \caption{Single Stage}
\label{fig:onestage1} 
\end{subfigure}
\begin{subfigure}[t]{0.23\textwidth}
  \centering
  \includegraphics[width=1.0\linewidth]{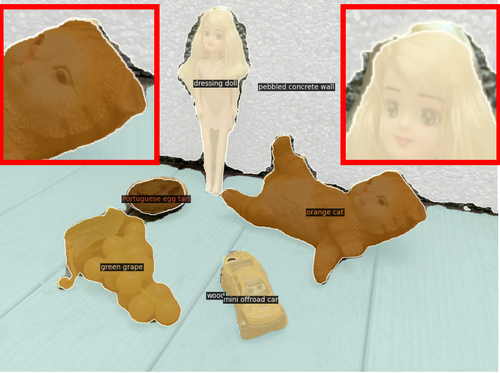}
  \caption{Two Stage} 
\label{fig:onestage2} 
\end{subfigure}
\caption{Qualitative comparison on the open vocabulary segmentation task between a Single Stage and Two Stage feature correction pipeline. }
\label{fig:baseline2}
\end{figure}

A straightforward extension of ~\cite{ye2023featurenerf}, that directly predicts the target view feature cannot generalize to unseen feature encoders resulting in a very high error rate. 
Instead, our method uses a feature interpolation strategy that derives consistency information from RGB to rectify and propagate inconsistent feature maps. 
We verify that our inconsistency correction module on the source view features $\{\Mat{G}_{i}\}_{i=1}^{N}$ is essential and ensures better blending of feature maps. 
This becomes even more apparent in the case of severely view-inconsistent input features, \eg in the case of the colorization task (DDColor~\cite{kang2022ddcolor}). 
Finally, our method is two-stage \ie derives a coarse density proposal and performs corrective feature aggregation only on the importance-sampled points. This is necessary and leads to slight deviations in the estimated novel view feature otherwise (and even worse decoded outputs, see Fig. \ref{fig:baseline2}).  

\section{Computational Efficiency}

Naively applying a 2D vision model on each rendered view yields multi-view inconsistent predictions and can be quite inefficient. 
On the other hand, our method {\em Lift3D} uses the 2D vision model to encode only the training views (can be pre-computed) and relies on nearest source views when estimating the features from arbitrary viewing angles, that are further decoded to obtain the desired output. This is significantly efficient and faster especially when performing the downstream task from a large number of arbitrary viewpoints.  

Formally, let's assume the time to encode, decode, and render each view as $t_{\text{enc}}$, $t_{\text{dec}}$, $t_{\text{rend}}$ respectively. To perform the desired task on 100 rendered views, the 2D baseline would roughly take time $t_{\text{2D}} = 100 \times (t_{\text{rend}} + t_{\text{enc}} + t_{\text{dec}})$. Assuming we have around 15 training views for each scene, the time taken by Lift3D $t_{\text{3D}} = 15 \times (t_{\text{enc}}) + 100 \times (t_{\text{rend}} + t_{\text{dec}})$. We can clearly see that $t_{\text{3D}} < t_{\text{2D}}$, and the difference is quite significant in the case of lifting diffusion features like InstructPix2Pix~\cite{brooks2023instructpix2pix} that requires a time-consuming multistep denoising process during encoding {\ie $t_{\text{enc}} \gg t_{\text{dec}}$}. Therefore, when performing downstream tasks on several arbitrary viewing angles, our method also boasts of superior efficiency along with multi-view consistency when compared to its corresponding 2D counterpart. 

\section{Limitations}

\begin{table}
  \centering
  \resizebox{\columnwidth}{!}{
  \begin{tabular}{lccc}
    \toprule
    2D DINO & \multicolumn{3}{c}{3D "Lifted" Dino}\\
    \midrule
    N.A. & 3 views & 6 views & 10 views\\
    \midrule
    0.39 / 0.90 / 0.36 & 0.76 / 0.97 / 0.76 & 0.80 / 0.98 / 0.80 & 0.82 / 0.98 / 0.83\\
    \bottomrule
  \end{tabular}}
    \caption{Effect of the number of input views for semantic segmentation of scenes. Metrics are ordered as IoU / Acc / mAP (higher is better)}
  \label{tab:ablviews}
\end{table}

We acknowledge that an interpolation technique like ours from input views does result in some loss in quality, arguably not very significant. 
However, in several practical applications multi-view consistent outputs are usually even more desirable, and even mild deviations from the current viewpoint yield significantly different outputs when naively applying a 2D vision model (see Figures \ref{fig:teaser}, \ref{fig:scene_edit}, \ref{fig:other_tasks}, \ref{fig:gallery}).
Although our method successfully lifts many 2D features to be multi-view consistent, its potential remains capped by the epipolar-based rendering. 
For example, our method may not handle sparse 360-degree scenes or objects with complex light transport where epipolar geometry no longer holds and drops in performance with limited number of input views (see Table \ref{tab:ablviews}).
Interesting future directions include scaling up training of Lift3D to unbounded scenes \cite{cong2023enhancing} or combining extant pre-trained 3D models with 2D models.

\section{Gallery of Tasks}

\begin{table}
  \centering
  \resizebox{0.99\columnwidth}{!}{
  \begin{tabular}{lcc}
    \toprule
    Method & Text-Image Direction Similarity $\uparrow$ & Direction Consistency $\uparrow$\\
    \midrule
    InstructNeRF2NeRF~\cite{haque2023instruct} & 0.180 & 0.966 \\
    Ours & \cellcolor{black!30}0.193 & \cellcolor{black!30}0.982 \\
    \bottomrule
  \end{tabular}}
  \caption{Quantitative results for text-guided scene editing.}
  \label{tab:sceneedit}
\end{table}

\begin{figure*}[t]
\centering
\raisebox{0.25in}{\rotatebox[origin=t]{90}{Stroke-Guided}}%
\raisebox{0.25in}{\rotatebox[origin=t]{90}{Segmentation}}%
\hspace{0.1em}
\begin{subfigure}[t]{0.19\textwidth}
  \centering
  \includegraphics[width=0.98\linewidth]{imgs/chess_table_stroke.png}
  \caption{User Stroke}
\label{fig:chesstable_dino_inp} 
\end{subfigure}%
\begin{subfigure}[t]{0.38\textwidth}
  \centering
  \includegraphics[width=0.49\linewidth]{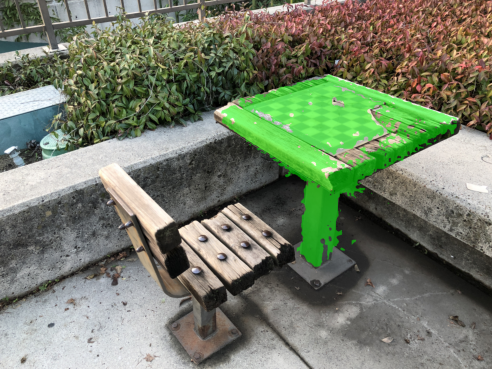}
  \includegraphics[width=0.49\linewidth]{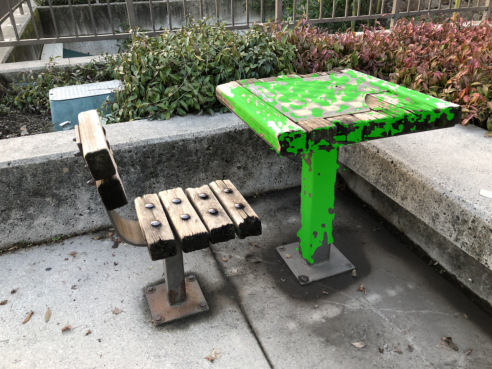}
  \caption{2D DINO}
\label{fig:gallery_dino_2d} 
\end{subfigure}%
\hspace{0.2em}
\begin{subfigure}[t]{0.38\textwidth}
  \centering
  \includegraphics[width=0.49\linewidth]{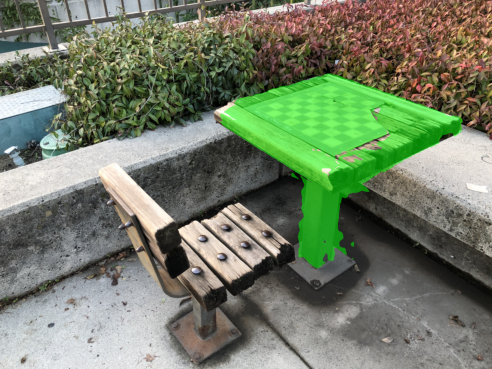}
  \includegraphics[width=0.49\linewidth]{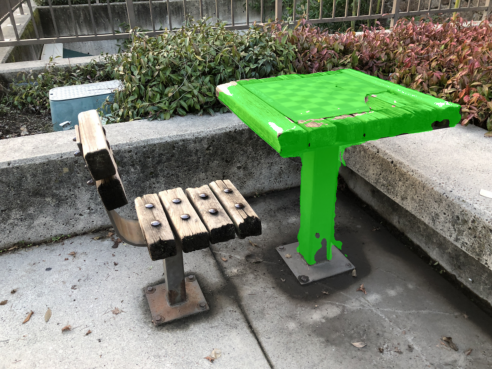}
  \caption{3D ``Lifted" DINO}
\label{fig:gallery_dino_3d} 
\end{subfigure}%
\newline
\raisebox{0.25in}{\rotatebox[origin=t]{90}{Text-Driven}}%
\raisebox{0.25in}{\rotatebox[origin=t]{90}{Scene Editing}}%
\hspace{0.1em}
\begin{subfigure}[t]{0.19\textwidth}
  \centering
  \includegraphics[width=0.98\linewidth]{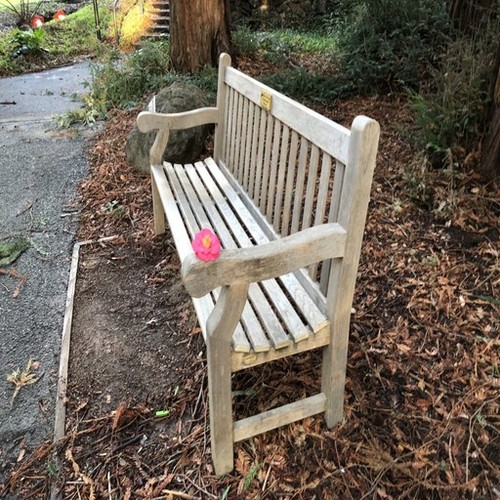}
  \caption{Original Scene \\ ``engulf the bench in \textcolor{red}{fire}"}
\label{fig:gallery_ip2p_inp} 
\end{subfigure}%
\begin{subfigure}[t]{0.38\textwidth}
  \centering
  \includegraphics[width=0.49\linewidth]{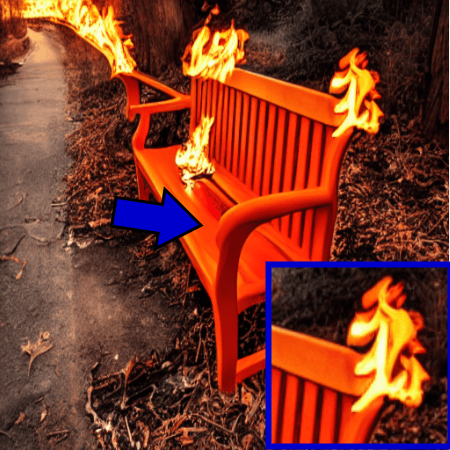}
  \includegraphics[width=0.49\linewidth]{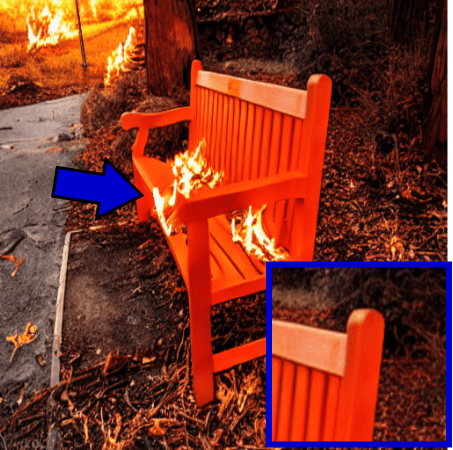}
  \caption{2D InstructPix2Pix}
\label{fig:gallery_ip2p_2d} 
\end{subfigure}%
\hspace{0.2em}
\begin{subfigure}[t]{0.38\textwidth}
  \centering
  \includegraphics[width=0.49\linewidth]{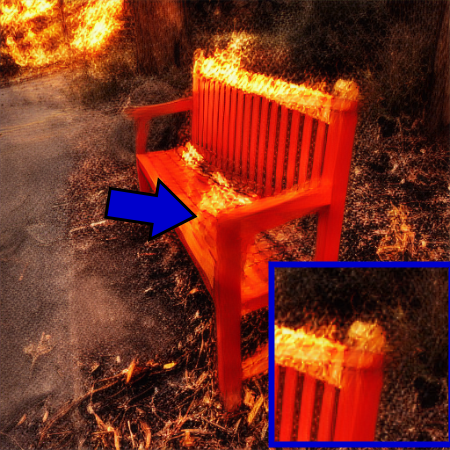}
  \includegraphics[width=0.49\linewidth]{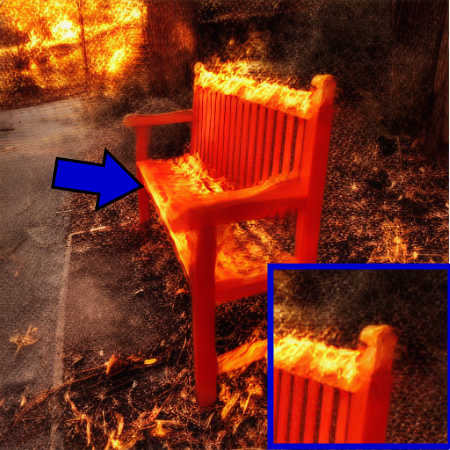}
  \caption{3D ``Lifted" InstructPix2Pix}
\label{fig:gallery_ip2p_3d} 
\end{subfigure}%
\newline
\raisebox{0.25in}{\rotatebox[origin=t]{90}{Colorization}}%
\hspace{0.1em}
\begin{subfigure}[t]{0.19\textwidth}
  \centering
  \includegraphics[width=0.98\linewidth]{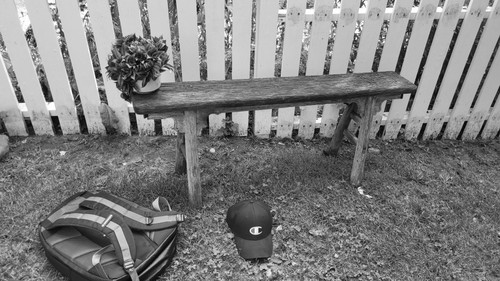}
  \caption{Gray Image}
\label{fig:gallery_ddcolor_inp} 
\end{subfigure}%
\begin{subfigure}[t]{0.38\textwidth}
  \centering
  \includegraphics[width=0.49\linewidth]{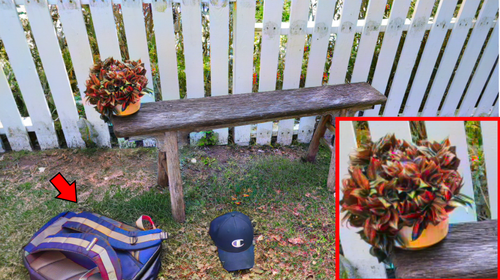}
  \includegraphics[width=0.49\linewidth]{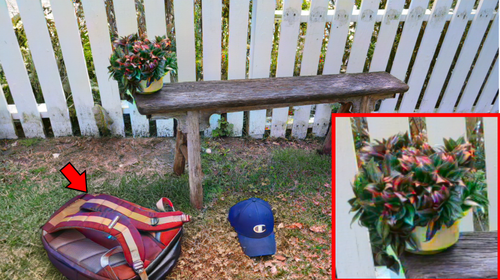}
  \caption{2D DDColor}
\label{fig:gallery_ddcolor_2d} 
\end{subfigure}%
\hspace{0.2em}
\begin{subfigure}[t]{0.38\textwidth}
  \centering
  \includegraphics[width=0.49\linewidth]{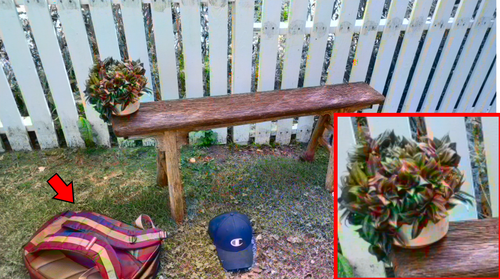}
  \includegraphics[width=0.49\linewidth]{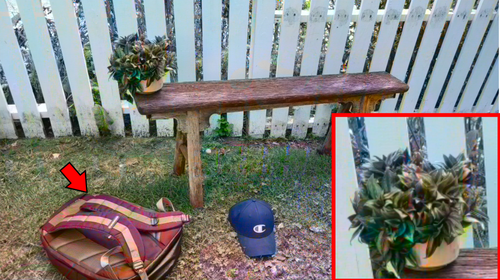}
  \caption{3D ``Lifted" DDColor}
\label{fig:gallery_ddcolor_3d} 
\end{subfigure}%
\newline
\raisebox{0.25in}{\rotatebox[origin=t]{90}{Open-Vocabulary}}%
\raisebox{0.25in}{\rotatebox[origin=t]{90}{Segmentation}}%
\begin{subfigure}[t]{0.19\textwidth}
  \centering
  \includegraphics[width=0.98\linewidth]{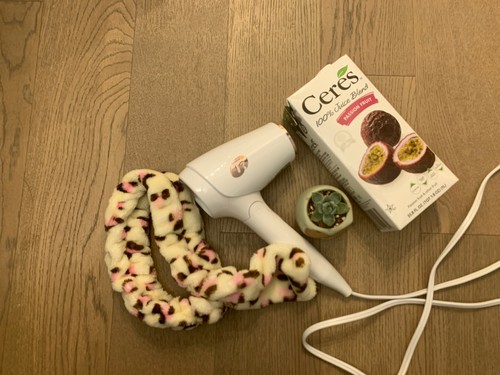}
  \caption{Original Scene}
\label{fig:gallery_ovseg_inp} 
\end{subfigure}%
\begin{subfigure}[t]{0.38\textwidth}
  \centering
  \includegraphics[width=0.49\linewidth]{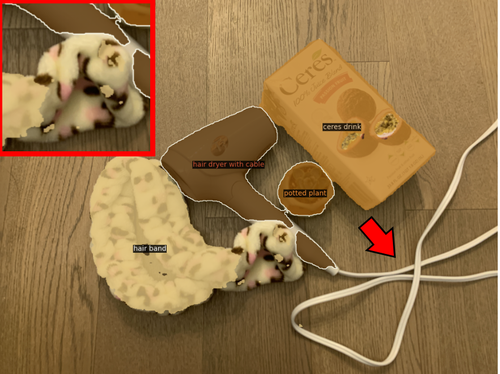}
  \includegraphics[width=0.49\linewidth]{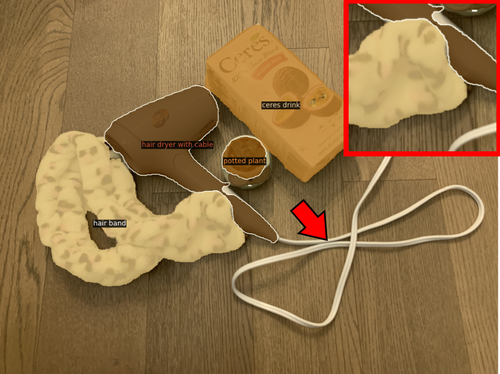}
  \caption{2D OVSeg}
\label{fig:gallery_ovseg_2d} 
\end{subfigure}%
\hspace{0.2em}
\begin{subfigure}[t]{0.38\textwidth}
  \centering
  \includegraphics[width=0.49\linewidth]{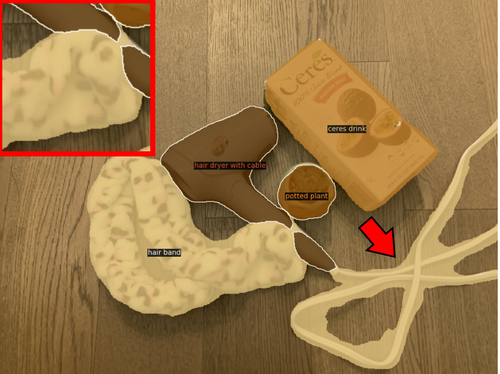}
  \includegraphics[width=0.49\linewidth]{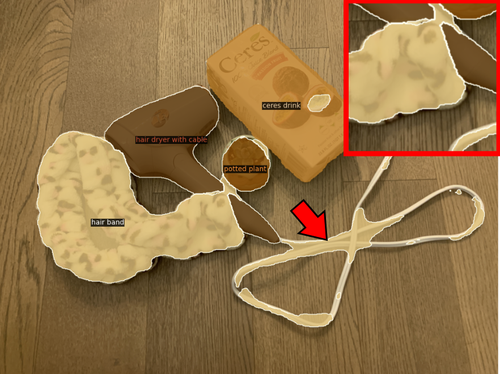}
  \caption{3D ``Lifted" OVSeg}
\label{fig:gallery_ovseg_3d} 
\end{subfigure}%
\newline
\raisebox{0.25in}{\rotatebox[origin=t]{90}{Super Resolution}}%
\hspace{0.1em}
\begin{subfigure}[t]{0.19\textwidth}
  \centering
  \includegraphics[width=0.98\linewidth]{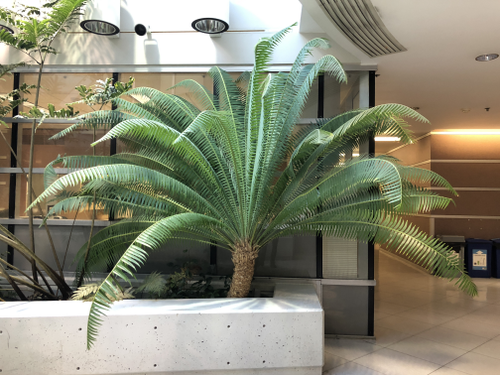}
  \caption{Low-Resolution Image (1/4x)}
\label{fig:gallery_resshift_inp} 
\end{subfigure}%
\begin{subfigure}[t]{0.38\textwidth}
  \centering
  \includegraphics[width=0.49\linewidth]{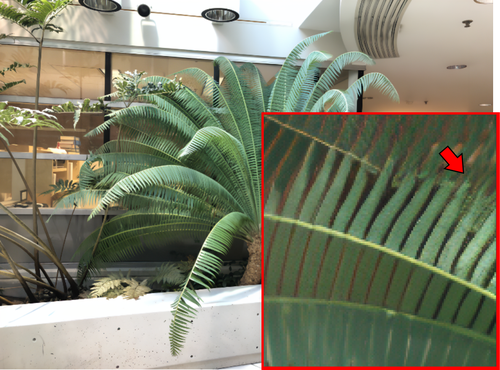}
  \includegraphics[width=0.49\linewidth]{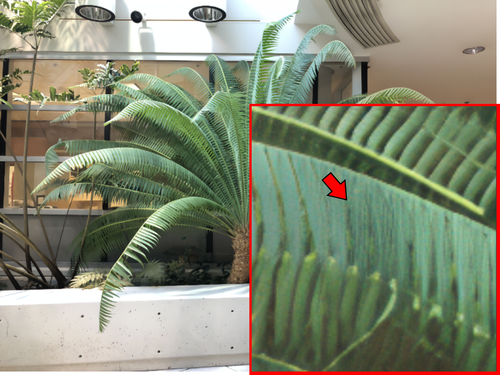}
  \caption{2D ResShift}
\label{fig:gallery_resshift_2d} 
\end{subfigure}%
\hspace{0.2em}
\begin{subfigure}[t]{0.38\textwidth}
  \centering
  \includegraphics[width=0.49\linewidth]{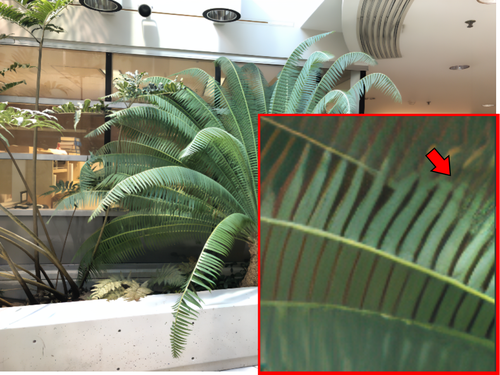}
  \includegraphics[width=0.49\linewidth]{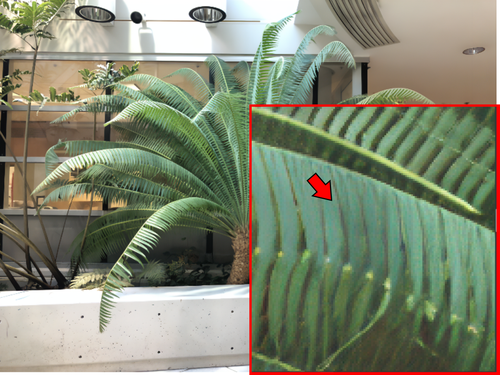}
  \caption{3D ``Lifted" ResShift}
\label{fig:gallery_resshift_3d} 
\end{subfigure}%
\caption{Qualitative comparisons of the 3D ``Lifted" features against its corresponding 2D counterpart on two different views and across several tasks. We observe that our 3D-corrected features are more multi-view consistent and sometimes even improve prediction quality. For clearer comparison between the 2D and 3D outcomes, we recommend zooming into the electronic version of this image. }
\label{fig:gallery}
\end{figure*}

In Fig. \ref{fig:gallery}, we qualitatively compare the decoded outputs using our 3D lifted features against the original 2D operation on two views and across different tasks. We can clearly see that our method yields multi-view consistent predictions, unlike the 2D operator. In some cases, we even see that our method yields improved predictions perhaps due to multi-view information. For example in Figs. \ref{fig:gallery_ovseg_2d} and \ref{fig:gallery_ovseg_2d}, we can see that our method is able to segment the hair dryer along with its cable as per the input prompt \textit{``hair dryer with cable''}. Similarly in Figs. \ref{fig:gallery_resshift_2d} and \ref{fig:gallery_resshift_3d}, in the case of super-resolution, we see that our lifted features preserve the original scene geometry with higher detail, unlike its 2D counterpart. For the sake of completeness, we also provide quantitative results for 3D scene editing in Table \ref{tab:sceneedit}. Following the evaluation protocol in ~\cite{haque2023instruct}, we compute the text-image direction similarity and consistency scores across 6 scenes and report average metrics. Our method outperforms the SOTA scene-specific 3D editing technique InstructNeRF2NeRF~\cite{haque2023instruct}, both in terms of editing quality and multi-view consistency. 
Fig. \ref{fig:gallery} only represents a few tasks and in practice, our method can be extended to any 2D vision operator without any extra tuning.

\end{document}